\documentclass[preprint,12pt]{elsarticle}

\usepackage{graphicx}
\usepackage{subfigure} 

\usepackage[justification=centering]{caption}

\usepackage{hyperref}
\usepackage{marvosym}

\pdfoutput=1

\usepackage{amssymb}

\usepackage{lineno}

\usepackage{url}

\biboptions{sort&compress}

\usepackage{geometry}
\usepackage{indentfirst}

\usepackage{amsmath}

\journal{\ }

\everymath{\displaystyle}

\begin{document}

\begin{frontmatter}



\title {Scaling transition from momentum stochastic gradient descent to plain stochastic gradient descent}


\author[a]{Kun Zeng}
\ead{zki@163.com}
\author[b]{Jinlan Liu}
\ead{liujl627@nenu.edu.cn}
\author[a]{Zhixia Jiang \corref{mycorrespondingauthor}} 
\author[b]{Dongpo Xu}
\ead{xudp100@nenu.edu.cn}
\cortext[mycorrespondingauthor]{Corresponding author: Zhixia Jiang( zhixia\_jiang@126.com) }

\address[a]{Department of Mathematics, Changchun University of Science and Technology, Changchun 130022, China}
\address[b]{School of Mathematics and Statistics, Northeast Normal University, Changchun 130024, China}

\begin{abstract}
The plain stochastic gradient descent and momentum stochastic gradient descent have extremely wide applications in deep learning due to their simple settings and low computational complexity. The momentum stochastic gradient descent uses the accumulated gradient as the updated direction of the current parameters, which has a faster training speed. Because the direction of the plain stochastic gradient descent has not been corrected by the accumulated gradient. For the parameters that currently need to be updated, it is the optimal direction, and its update is more accurate. We combine the advantages of the momentum stochastic gradient descent with fast training speed and the plain stochastic gradient descent with high accuracy, and propose a scaling transition from momentum stochastic gradient descent to plain stochastic gradient descent(TSGD) method. At the same time, a learning rate that decreases linearly with the iterations is used instead of a constant learning rate. The TSGD algorithm has a larger step size in the early stage to speed up the training, and training with a smaller step size in the later stage can steadily converge. Our experimental results show that the TSGD algorithm has faster training speed, higher accuracy and better stability. Our implementation is available at: \href{https://github.com/kunzeng/TSGD}{https://github.com/kunzeng/TSGD}.

\end{abstract}


\end{frontmatter}


\section{Introduction}
\label{S:1}

In recent years, with the development of deep learning technologies, neural networks have been extremely widely used\cite{ma2021research,wang2021review,farooq2021advances}. Gradient descent is an important algorithm for training neural networks. As an unconstrained optimization method, the stochastic gradient descent is applied in many fields, especially in neural networks. However, the stochastic gradient descent has some disadvantages, such as slow training speed and unstable training. Researchers have proposed many adaptive gradient descent methods and their variants, such as AdaGrad\cite{duchi2011adaptive}, Adam\cite{kingma2014adam}, AmsGrad\cite{reddi2019convergence} and so on. However, the adaptive gradient descent needs to calculate the second moment, which has high computational complexity. In some aspects, the generalization ability of adaptive gradient descent is often not as good as stochastic gradient descent\cite{wilson2017marginal}. In practical applications, most neural networks still use stochastic gradient descent for training, such as ResNet\cite{he2016deep}, DensNet\cite{huang2017densely} and so on.

Plain stochastic gradient descent\cite{robbins1951stochastic} uses the negative direction of the gradient of loss function as the  direction of parameters update. The hyperparameter setting is relatively simple and has low computational complexity. It has higher efficiency for large-scale data training. Polyak et al.\cite{polyak1964some} improved the plain stochastic gradient descent and proposed a heavy-ball momentum stochastic gradient descent. The heavy-ball momentum stochastic gradient descent uses the accumulated historical gradient as momentum. Then it uses momentum to update the current parameters. So as to speed up training and has better stability. Nesterov's accelerated gradient (NAG)\cite{2019Reducing} is a modification of the momentum-based update which uses a look-ahead step to improve the momentum term\cite{1990Handwritten}. These two momentum gradient descent methods have a wide range of applications due to their simple settings and good generalization ability. Subsequently, researchers improved the momentum stochastic gradient descent, and proposed synthesized Nesterov variants\cite{lessard2016analysis}, PID control\cite{BenRecht2018Thebest}, AccSGD\cite{kidambi2018insufficiency}, SGDP\cite{heo2021adamp} and so on.

In 2019, Jerry Ma et al.\cite{ma2018quasi} proposed the quasi-hyperbolic momentum algorithm. For the direction of parameters update, QHM algorithm mainly considers a linear combination of the last momentum and current gradient, rather than just using momentum to update the parameters. The more important thing of this algorithm is a general structure of the momentum stochastic gradient descent. Through the adjustment of the parameters $v$ and $\beta $, it can be transformed into the Nesterov’s accelerated gradient\cite{2019Reducing}, PID control\cite{BenRecht2018Thebest}, synthesized Nesterov variant\cite{lessard2016analysis}, least-squares acceleration of SGD algorithm\cite{kidambi2018insufficiency}. 

We can see that the historical gradient is used in the momentum stochastic gradient descent to make a correction to the current gradient direction, thereby speeding up the training. In fact, we can empirically see that momentum is an exponentially weighted combination of historical gradient and current gradient. Since the past gradient is not computed based on the most updated stochastic gradient descent state, it can introduce a deviation to the new gradient computation, negatively impacting its rate of convergence\cite{song2021ag}. Its update direction is no longer the optimal direction for the current parameters, which affects the accuracy. For the update direction of plain stochastic gradient descent each time, it is the optimal direction of the current parameters. Although it may be easy to fall into the local optimum, the training speed is slow, etc. But for the current parameters, the plain stochastic gradient descent has a more accurate direction. Therefore, for the gradient descent, how to get fast training speed and high accuracy is very important.

In this paper, we combine the fast training speed of the heavy-ball momentum stochastic gradient descent and the high accuracy of the plain stochastic gradient descent. By scaling, we achieve a smooth and stable transition from momentum stochastic gradient descent to plain stochastic gradient descent. In the early stage of training, we use momentum stochastic gradient descent, and in the later stage we use plain stochastic gradient descent for training. It has a faster training speed in the early stage, and a higher accuracy in the later stage of training. And this transition process is automatic, smooth and stable. The experimental results show that our proposed algorithm has a faster training speed and better accuracy in the neural network model.

\section{Preliminaries}
\label{S:2}

\textbf {\fontsize{12pt}{0} \selectfont Optimization problem.} We consider the following unconstrained convex optimization problem\cite{savarese2019convergence}:
\[\mathop {\min }\limits_{\theta \in X} {\rm{ }}F\left( \theta  \right).\]
Where $F$ is a continuous differentiable convex function on $X$. We use the gradient descent to solve the above problem

\textbf {\fontsize{12pt}{0} \selectfont Plain stochastic gradient descent.} In the plain stochastic gradient descent, we randomly select the sample  and use the current gradient to update the parameters\cite{loizou2020momentum}:
\[{\theta _{t + 1}} = {\theta _t} - {\eta}{g_t}.\]
Where $\theta $ is the parameter that needs to be optimized, and ${g_t}$ is the gradient of the current loss function. ${\eta}$ is the step size.

\textbf {\fontsize{12pt}{0} \selectfont Momentum stochastic gradient descent.} The momentum stochastic gradient descent considers the historical gradient information, and uses it to make a correction to the direction of the current gradient, which speeds up the training. The update formula of the heavy-ball momentum stochastic gradient descent is\cite{ghadimi2015global}:
$$
\begin{array}{l}
m_{t} = \beta m_{t-1}+(1-dampening) g_{t}, \\
\theta_{t+1} = \theta_{t}-\eta_{t} \hat{m}_{t}.
\end{array}
$$
Where ${m_t}$ is momentum, $\beta $ is momentum factor, and $dampening$ is the dampening for momentum$(dampening=0)$.

\textbf {\fontsize{12pt}{0} \selectfont Momentum and gradient.} In the heavy-ball momentum stochastic gradient descent, the accumulation of historical gradient is used as the momentum. It uses the momentum direction to replace the current gradient direction for parameters update. Although the momentum stochastic gradient descent reduces the variance, it has an acceleration effect. But from the composition of the momentum term, we can empirically see that the negative direction of the gradient of the loss function $\left( { - \nabla {F_t}\left( {{\theta _t}} \right)} \right)$ is the optimal direction for the current parameters update. If the correction is made to the direction of the current gradient by using the momentum, the momentum direction is no longer the optimal direction for the current parameters. 

\begin{figure}[htbp]
\centering
\subfigure[The accuracy on test set]{
\begin{minipage}[t]{0.5\linewidth}
\centering
\includegraphics[width=8cm]{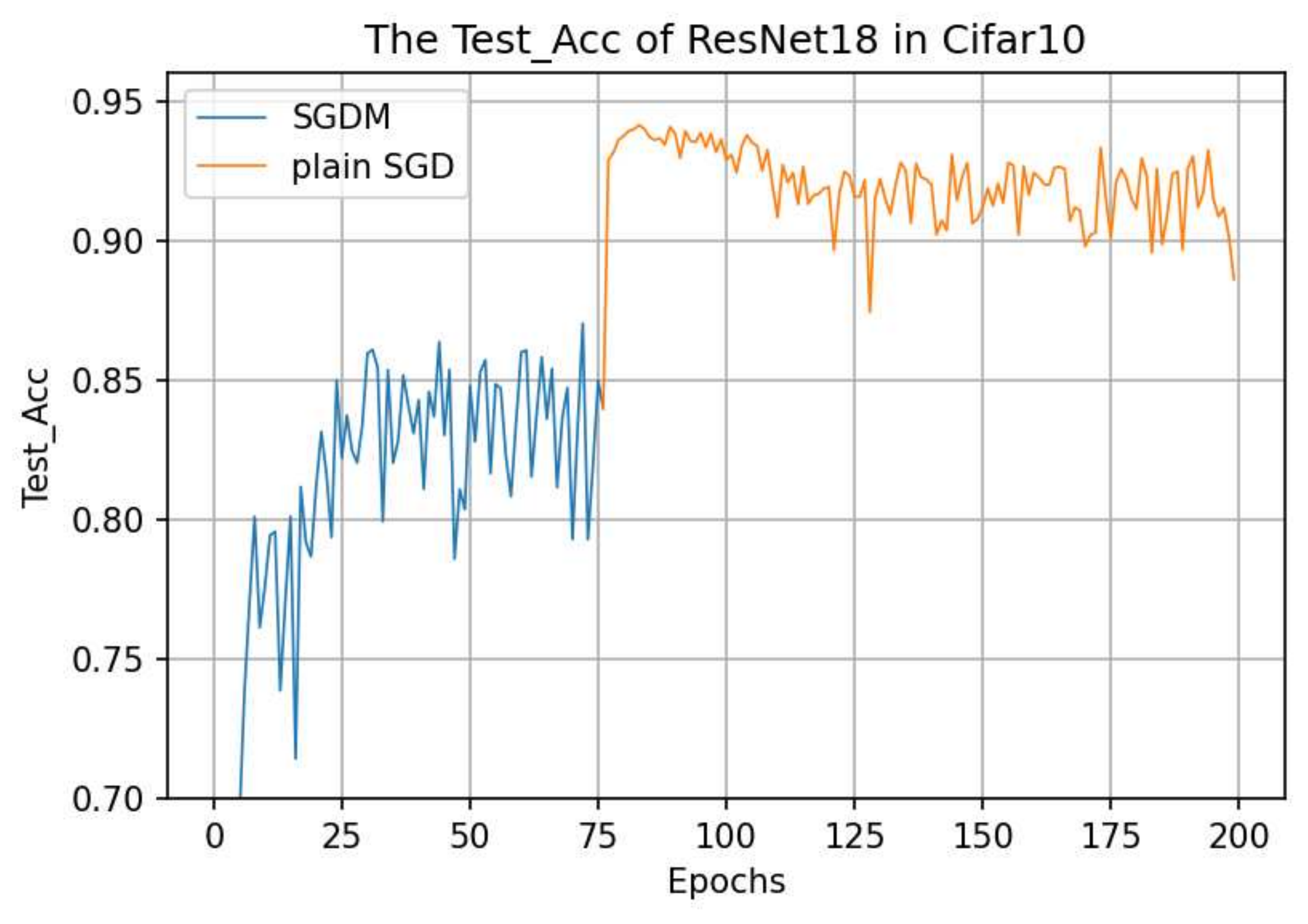}
\hspace{0.1in}
\end{minipage}%
}%
\subfigure[The loss on training set]{
\begin{minipage}[t]{0.5\linewidth}
\centering
\includegraphics[width=8cm]{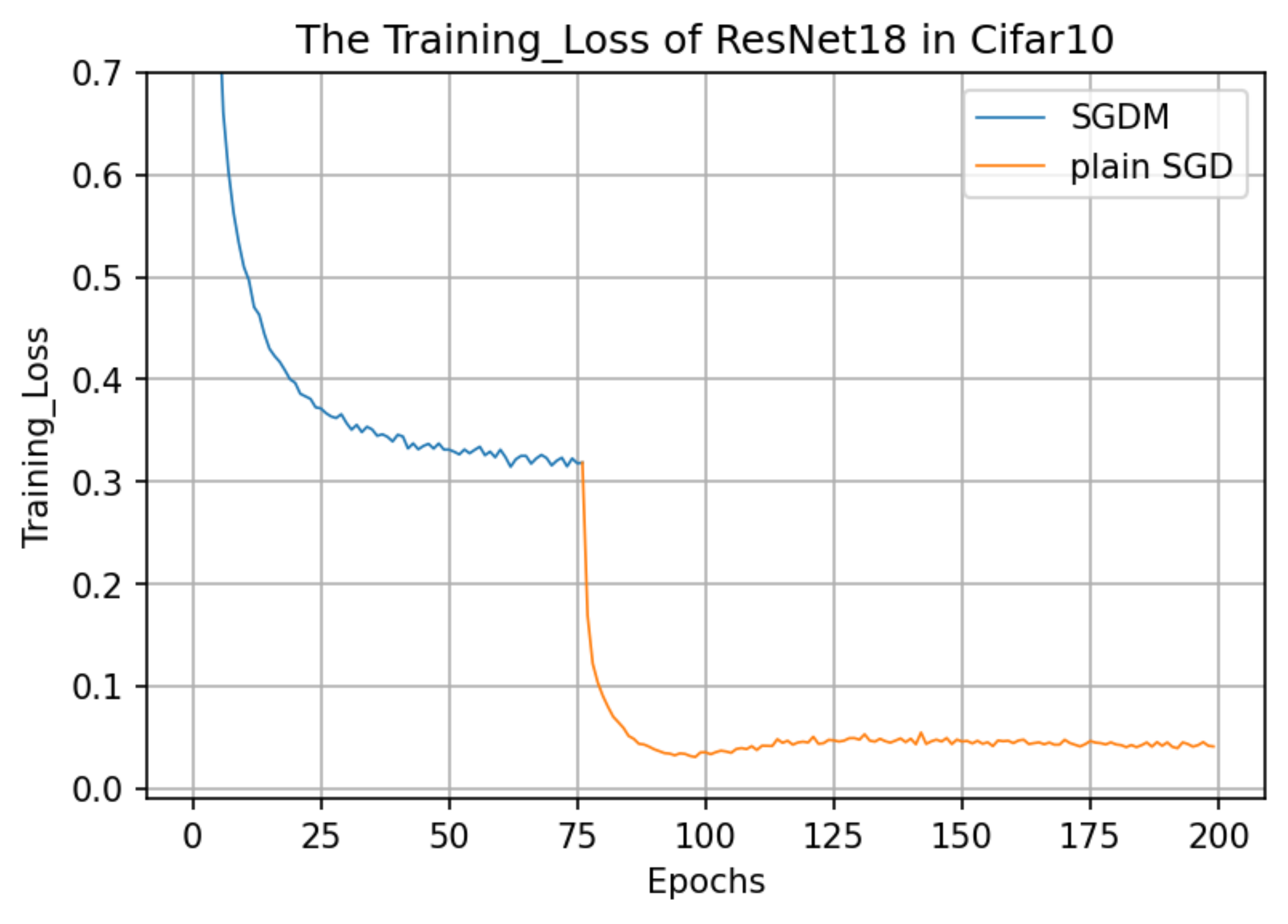}
\hspace{0.1in}
\end{minipage}%
}%
\caption{We use ResNet18 to train cifar10 data set, the first 75 epochs use momentum stochastic gradient descent(SGDM), and the remaining epochs use plain stochastic gradient descent(plain SGD) for training}
\end{figure}

We use the ResNet18 to train the cifar10 data set, and use the SGDM algorithm training for the first 75 iterations (Hyperparameters settings: weight\_decay=$5e-4$, $lr$=$0.1$, $\beta$=$0.9$, dampening=0). After the 75th iteration, we change the momentum to the current gradient (if epoch $<$ 76:  ${m_t}$ else: ${g_t}$, the learning rate has not been changed, always $lr$=$0.1$). Picture (a) of Figure 1 shows the accuracy of the test set during the training process. It can be seen that after the accuracy increases rapidly from starting, the SGDM algorithm enters the plateau after about 75-$th$ iteration, and the accuracy no longer increases. After 75-$th$ iteration, the current gradient direction is used to update the parameters, and it can be seen that the accuracy has been significantly improved. This also shows from the side that the current negative gradient direction is  more accurate. Picture (b) of Figure 1 shows the loss value of the training set. It can be seen that the loss of the plain SGD method is faster and smaller. But for deep learning, a faster training speed is often very important. Therefore, a good idea is to combine fast speed and high accuracy for the gradient descent method.

\textbf {\fontsize{12pt}{0} \selectfont Assumption A.} The following conditions hold for $F$  and the stochastic gradient oracle\cite{gitman2019understanding}:

1. $F$ is differentiable and $\nabla F$ is Lipschitz continuous, i.e., there is a constant $L$ such that
\[\left\| {\nabla F\left( x \right) - \nabla F\left( y \right)} \right\| \le L\left\| {x - y} \right\|,x,y \in {R^n}.\]

2. $F$ is bounded below and $\left\| {\nabla F\left( x \right)} \right\|$ is bounded above, i.e., there exist  ${F_*}$ and  $G$ such that
\[F\left( x \right) \ge {F_*},\left\| {\nabla F\left( x \right)} \right\| \le G,x \in {R^n}.\]

3. For $k = 0,1,2, \ldots ,$ the stochastic gradient ${g^k} = \nabla F\left( {{x^k}} \right) + {\xi ^k}$  , where the random noise ${\xi ^k}$  satisfies
\[{E_k}\left[ {{\xi ^k}} \right] = 0,{E_k}\left[ {{{\left\| {{\xi ^k}} \right\|}^2}} \right] \le C.\]
where  ${E_k}\left[  \cdot  \right]$ denotes expectation conditioned on  $\left\{ {{x^0},{g^0}, \ldots ,{x^{k - 1}},{g^{k - 1}},{x^k}} \right\}$, and $C$  is a constant.

\textbf {\fontsize{12pt}{0} \selectfont Lemma 1.}  Let $F$  satisfy Assumption A. Additionally, assume  $0 \le {v_k} \le 1$ and the sequences  $\left\{ {{a_k}} \right\}$
and $\left\{ {{\beta _k}} \right\}$  satisfy the following conditions\cite{gitman2019understanding}:
\[\sum\limits_{k = 0}^\infty  {{\alpha _k}}  = \infty ,\sum\limits_{k = 0}^\infty  {\alpha _k^2}  = \infty ,\mathop {\lim }\limits_{k \to \infty } {\beta _k} = 0,\bar \beta  \buildrel \Delta \over = \mathop {\sup }\limits_k {\beta _k} < 1.\]
Then the sequence  $\left\{ {{x_k}} \right\}$ generated by the QHM in literature\cite{gitman2019understanding}. Moreover, we have:
$$
\limsup _{k \rightarrow \infty} F\left(x^{k}\right)=\limsup _{k \rightarrow \infty,\left\|\nabla F\left(x^{k}\right)\right\| \rightarrow 0} F\left(x^{k}\right).
$$

\section{TSGD Algorithm}
\label{S:3}

Based on the above analysis and the basis of the QHM algorithm, we combine the advantages of the momentum stochastic gradient descent with fast training speed and the plain stochastic gradient descent with high accuracy. The scaling transition from momentum stochastic gradient descent to plain stochastic gradient descent method is proposed. At the same time, a learning rate that decreases linearly with the number of iterations is used instead of a constant learning rate. It allows the momentum stochastic gradient descent with a larger step size to be used in the initial stage to speed up the training, and the plain stochastic gradient descent with a smaller step size for training in the later stage can steadily converge and avoid oscillation. The process of this transformation is realized with the scaling by the iterations, which is automatic, smooth and stable. TSGD algorithm is described as Algorithm 1.

\begin{table}[!ht]

\centering
\begin{tabular}{l l}
\hline
\textbf{Algorithm 1:} Scaling \textbf{T}ransition from Momentum Stochastic Gradient descent to Plain \\ \textbf{S}tochastic \textbf{G}radient \textbf{D}escent(TSGD). \\
\hline
\textbf{Input:} $x \in \mathcal{F}$, $UR:$ upper learning rate, $LR:$ lower learning rate, $\gamma $: scaling factor, ${\beta}$: \\ momentum factor, ${\eta_t}:$ learning rate, $T:$ epochs,  $dampening$: dampening for momentum.\\
\textbf{Initialize:} $m_{0}=0,t=0$ \\
\textbf{for} ${t}$ \textbf{to} ${T-1}$ \textbf{do}\\ 
\hspace{2em} $t = t + 1$\\
\hspace{2em} ${g_t} = \nabla {f_t}({\theta _t})$ \hspace{10em}\ \ \ Update the gradient of objective function\\
\hspace{2em} ${m_t} = {\beta}{m_{t - 1}} + (1 - dampening){g_t}$ \hspace{1em} \ Update first moment momentum\\
\hspace{2em} $\hat{m}_{t}=\left(m_{t}-g_{t}\right) \frac{1}{\gamma^{t}}+g_{t}$  \hspace{5em}\ \ \ \ \ Scaling the momentum and gradient\\
\hspace{2em} $\eta_{t}=(U R-L R)(1-t / T)+L R$ \hspace{1em}\ \ \ \ Decreasing learning rate\\
\hspace{2em} $\theta_{t+1} = \theta_{t}-\eta_{t} \hat{m}_{t}$ \hspace{8em}\ \ \ \ Update parameters\\
\textbf{end for}\\
\hline
\end{tabular}
\end{table}

Oracle recommends an empirical value of: $\beta = 0.9$, $UR = 0.1$, $LR  = 0.005$$\left( {UR \ge LR} \right)$, $\gamma $$\left( {\gamma  \ge 1} \right)$. Please see section 6 for details on the value of $UL$, $LR$, $\gamma $. ${g_t}$: gradient of loss function $f$ in $t$-th iteration. ${m_t}$: momentum. ${\hat m_t}$: scaled momentum. ${\theta _t}$: optimized parameters.

In Algorithm 1, we use a linearly decreasing learning rate with the number of iterations instead of a constant learning rate, give as:
\[{\eta _t} = \left( {UR - LR} \right)(1 - t/T) + LR.\]
From the above formula, the TSGD algorithm can have a larger step size in the early stage, which can speed up the training. In the later stage of training, having a small step size can avoid skipping the optimal solution or oscillating, and makes training more stable. 

For the transition process, we implemente the following formula:
\[{\hat m_t} = \left( {{m_t} - {g_t}} \right)\frac{1}{{{\gamma ^t}}} + {g_t}.\]
We rearrange the formula of parameters updated as follows:
\[{\theta _{t + 1}} = {\theta _t} - {\eta _t}\left[ {\frac{\beta }{{{\gamma ^t}}}{m_{t - 1}} + \left( {1 - \frac{dampening }{{{\gamma ^t}}}} \right){g_t}} \right].\]
It can be seen that the direction of the parameters update of the TSGD algorithm is a linear combination of the last momentum and the current gradient. But the weight coefficient here decreases as the iteration increasing. When $t$ is smaller, $1/{\gamma ^t}$ is close to 1, the TSGD algorithm is momentum stochastic gradient descent. When $t$ is larger, $1/{\gamma ^t}$ is close to 0, and the TSGD algorithm is plain stochastic gradient descent. As the iteration $t$ increasing, $1/{\gamma ^t}$ achieves a smooth and stable transition from the momentum stochastic gradient descent to the plain stochastic gradient descent.

The comparison of TSGD and QHM algorithm\cite{ma2018quasi} shows that QHM algorithm is a special case of TSGD algorithm. In the TSGD algorithm, when ${\gamma ^t}$ is $\gamma $, the TSGD algorithm is the QHM algorithm. In the literature\cite{ma2018quasi}, the author analyzes the relationship between QHM algorithm and Nesterov's accelerated gradient, PID control\cite{BenRecht2018Thebest}, synthesized Nesterov varian $\left( SNV \right)$\cite{lessard2016analysis}, AccSGD\cite{kidambi2018insufficiency} algorithms in detail. At the same time, in the deterministic (full-batch) case, QHM can recover the global linear convergence rate of $1 - 1/\sqrt k $ for strongly convex, smooth loss functions when QHM recovers Triple Momentum. In the stochastic (minibatch) case, QHM’s recovery of AccSGD gives QHM the same convergence results as in the literature\cite{kidambi2018insufficiency}'s least-squares regression setting, of ${\rm O}\left( {\sqrt \kappa   \cdot \log \kappa  \cdot \log \frac{1}{\epsilon }} \right)$ iterations for $\epsilon$-approximation of the minimal loss. The QHM algorithm is a special case of the TSGD algorithm. Obviously, the relationship and convergence between different algorithms and QHM in the literature\cite{ma2018quasi} have the same results as the TSGD algorithm. In particular, we take $\beta_{k} = \beta = dampening$, ${v_k} = 1/{\gamma ^k}$, under the Assumption A, we can get the same Lemma 1 of TSGD algorithm.

In fact, the idea of combining last momentum stochastic gradient and plain stochastic gradient is not the first time to appear in this paper. For example, the QHM\cite{ma2018quasi} algorithm uses the moving weighted average last momentum and current gradient as the current update direction. The main formula is:
$$
\begin{array}{l}
g_{t+1} \leftarrow \beta \cdot g_{t}+(1-\beta) \cdot \nabla \hat{L}_{t}\left(\theta_{t}\right), \\
\theta_{t+1} \leftarrow \theta_{t}-\alpha\left[(1-\nu) \cdot \nabla \hat{L}_{t}\left(\theta_{t}\right)+\nu \cdot g_{t+1}\right].
\end{array}
$$
The author gives an empirical value of $v = 0.7$ in the paper. In the literature\cite{arnold2019reducing}, the author transforms gradients computed at previous iterates into gradients evaluated at the current iterate without using the Hessian explicitly. The momentum factor is taken as a function with the iterations, and its main formula is:
$$
\begin{array}{l}
\gamma_{t} \leftarrow \frac{t}{t+1}, \\
v_{t} \leftarrow \gamma_{t} v_{t-1}+\left(1-\gamma_{t}\right) g\left(\theta_{t}+\frac{\gamma_{t}}{1-\gamma_{t}}\left(\theta_{t}-\theta_{t-1}\right), x_{t}\right).
\end{array}
$$

From the above, we can see that although many researchers have explored the combination of momentum stochastic gradient descent and plain stochastic gradient descent, and many achievements have been made. But according to the best of our knowledge, the transition from momentum stochastic gradient descent to plain stochastic gradient descent by scaling with the iterations is proposed in this paper for the first time. In our proposed TSGD algorithm, we use a linearly decreasing learning rate instead of the constant learning rate, so that the algorithm has a larger step size in the early stage to speed up, and a smaller step size in the later stage of training can be stable convergence. At the same time, we achieved a smooth and stable transition from momentum stochastic gradient descent to stochastic gradient descent through scaling. It combines the advantages of the momentum stochastic gradient descent with fast training speed and the plain stochastic gradient descent with high accuracy. Making the TSGD algorithm has a faster convergence speed and also a higher accuracy.

\section{TAdam Algorithm}
\label{S:4}

Plain stochastic gradient descent, due to the slow convergence speed and unstable training. It often needs to manually adjust the learning rate through the interval learning rate (StepLR). In 2016, Duchi et al.\cite{polyak1964some} proposed the adaptive gradient descent AdaGrad using the second moment as the adaptive learning rate. The adaptive gradient descent has a faster training speed and the training process is more stable. It is more suitable for different scenarios, and does not need or very few parameters need to be manually adjusted. Then researchers improved it, and Adam\cite{kingma2014adam} algorithm has good performance as an excellent representative of adaptive gradient descent. It also has a wide range of applications in the field of deep learning. We use the scaled momentum term in TSGD to replace the momentum term in Adam's algorithm. As a result, the descent direction in Adam's algorithm transform from the momentum stochastic gradient descent direction to the plain stochastic gradient descent direction with the iterations. We name it TAdam algorithm, such as Algorithm 2.

\begin{table}[!ht]

\centering
\begin{tabular}{l l}
\hline
\textbf{Algorithm 2:} \textbf{Adam} with Scaling \textbf{T}ransition from Momentum stochastic gradient \\ descent to Plain Stochastic Gradient Descent(TAdam). \\
\hline
\textbf{Input:} $x \in \mathcal{F}$, $\gamma $: scaling factor($\gamma \ge 1$,see section 6 for details), ${\beta _1},{\beta _2}:$ constants, ${\eta}:$ \\global learning rate, $T:$ epochs, $\varepsilon $: regularization constant.\\
\textbf{Initialize:} $m_{0}=0, v_{0}=0, t=0$ \\
\textbf{for} ${t}$ \textbf{to} ${T-1}$ \textbf{do}\\ 
\hspace{2em} $t = t + 1$\\
\hspace{2em} ${g_t} = \nabla {f_t}({\theta _t})$ \hspace{10em}\ \ \ Update the  gradient of objective function\\
\hspace{2em} ${m_t} = {\beta _1}{m_{t - 1}} + (1 - {\beta _1}){g_t}$ \hspace{5em} Update first moment\\
\hspace{2em} $\hat{m}_{t}=\left(m_{t}-g_{t}\right) \frac{1}{\gamma^{t}}+g_{t}$  \hspace{5em}\ \ \ \ \ Scaling the momentum and gradient\\
\hspace{2em} ${v_t} = {\beta _2}{v_{t - 1}} + (1 - {\beta _2})g_t^2$  \hspace{5em}\ \ \ Update second moment\\
\hspace{2em} ${\theta _{t + 1}} = {\theta _t} - \eta \frac{{{{\hat m}_t}}}{{\sqrt {{v_t}}  + \varepsilon }}$ \hspace{6em}\ \ \ \ Update parameters\\
\textbf{end for}\\
\hline
\end{tabular}
\end{table}

The authors of proposed Adam algorithm gave some empirical values ${\beta _1} = 0.9$, ${\beta _2} = 0.999$, $\eta  = 0.001$. ${f_t}\left( {{\theta _t}} \right)$: loss function,  ${g_t}$: gradient of loss function, ${m_t}$: momentum, ${\hat m_t}$: scaled momentum,  ${v_t}$: adaptive learning rate, ${\theta _t}$: optimized parameters. From Algorithm 2, it can be seen that when $\gamma  = 1$, Algorithm 2 is Adam algorithm, when $\gamma  > 1$, Algorithm 2 is TAdam algorithm.

\section{Experiments}
\label{S:5}

In this section, in order to verify the performance of the TSGD algorithm, we select different data sets, different neural network models and different training algorithms for comparison. The experimental environment is: CPU: Intel(R) Core(TM) i7-6500U CPU @ 2.5GHz 2.60GHz, 16 cores; memory: 32GB; operating system: Centos8.3; programming language: Python3.7 (Aanaconda3); deep learning framework: Pytorch1.7.0; graphics card: Quadro P600, CUDA 11.0, cuDNN v8.0.4; graphics card memory: 2GB. In order to ensure fairness, we set the same random number seed through pytorch, so that it is not affected by other factors except the training algorithm.

\textbf {\fontsize{12pt}{0} \selectfont MNIST and FCNN.} MNIST\cite{lecun1998gradient} is a classic data set for testing model performance in the field of deep learning. This data set consists of 0-9 handwritten digits, with a total of 70,000 pictures, of which 60,000 are the training samples and 10,000 are the test samples. We select the classic fully connected neural network(FCNN) for training. The structure of FCNN is: 2 hidden layers, each hidden layer has 512 neurons. The loss function is the cross-entropy function. When the batch is divisible by 100, we record a result of the experiment. The hyperparameters of the TSGD algorithm is set to: ($UR=0.1$, $LR=0.005$, $\beta=0.9$, $\gamma=1.000767$, epoch $=20$, dampening = $0$, batchSize $=100$, weight$_{-}$decay = 5$e$-4, seed = 1).

We chose SGDM, AdaGrad, Adam, AdaBound algorithms to compare with the proposed TSGD in this paper. Figure 2 shows the accuracy on the test set. It can be seen that the TSGD algorithm has a better performance on the test set, more stable, higher accuracy, faster speed. Figure 3 shows the loss on the training set. It can be seen that the loss of each algorithm has larger oscillation, but compared with SGDM, it can be seen that the TSGD algorithm has smaller loss, and the oscillation is also relatively small. We use the optimal value of the training process as the result of this experiment, and the results are shown in Table 1.

\begin{figure}[!h]
\centering
\begin{minipage}[t]{0.48\textwidth}
\centering
\includegraphics[width=8cm]{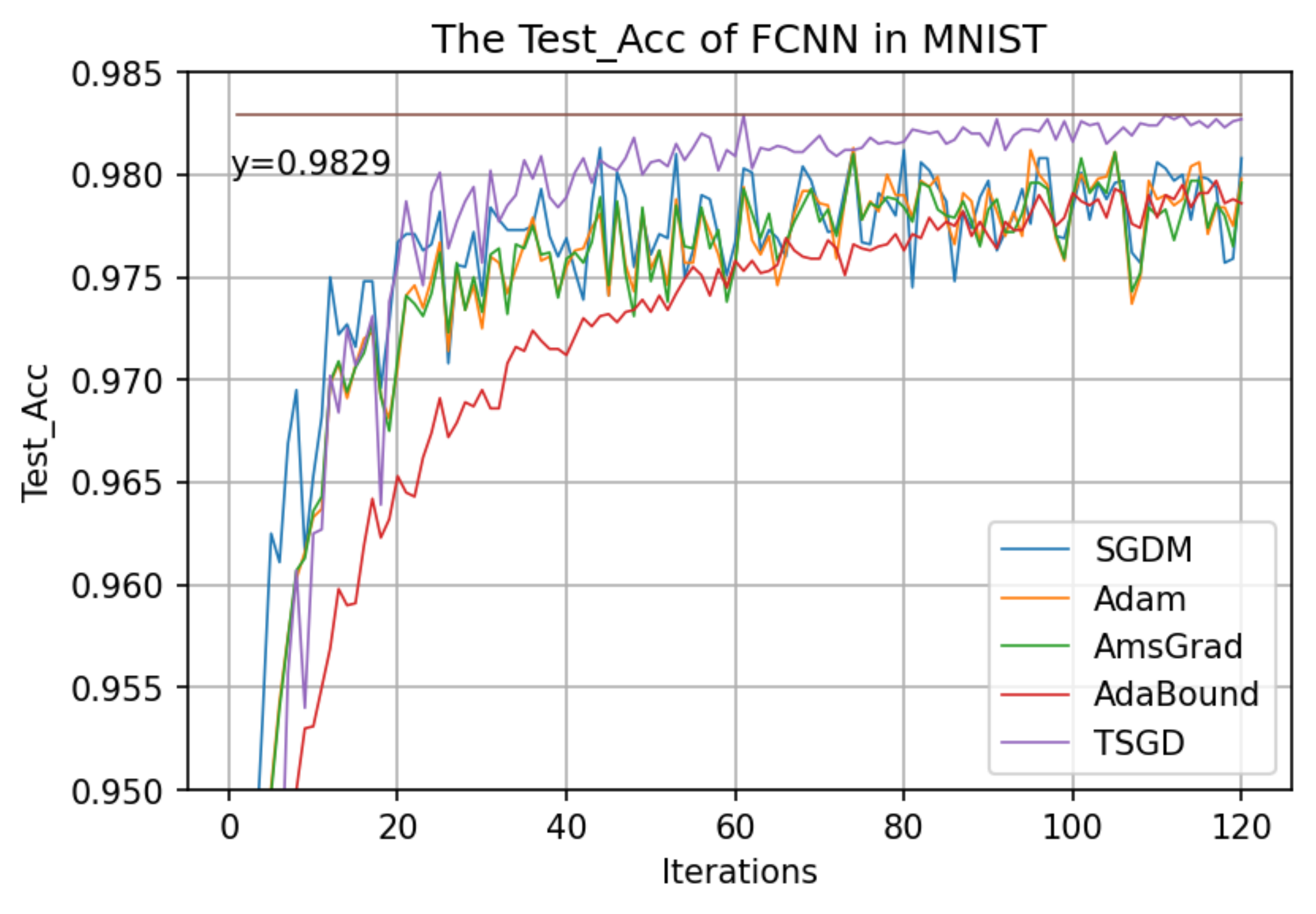}
\caption{The accuracy of MNIST-FCNN on test set}
\end{minipage}
\begin{minipage}[t]{0.48\textwidth}
\centering
\includegraphics[width=8cm]{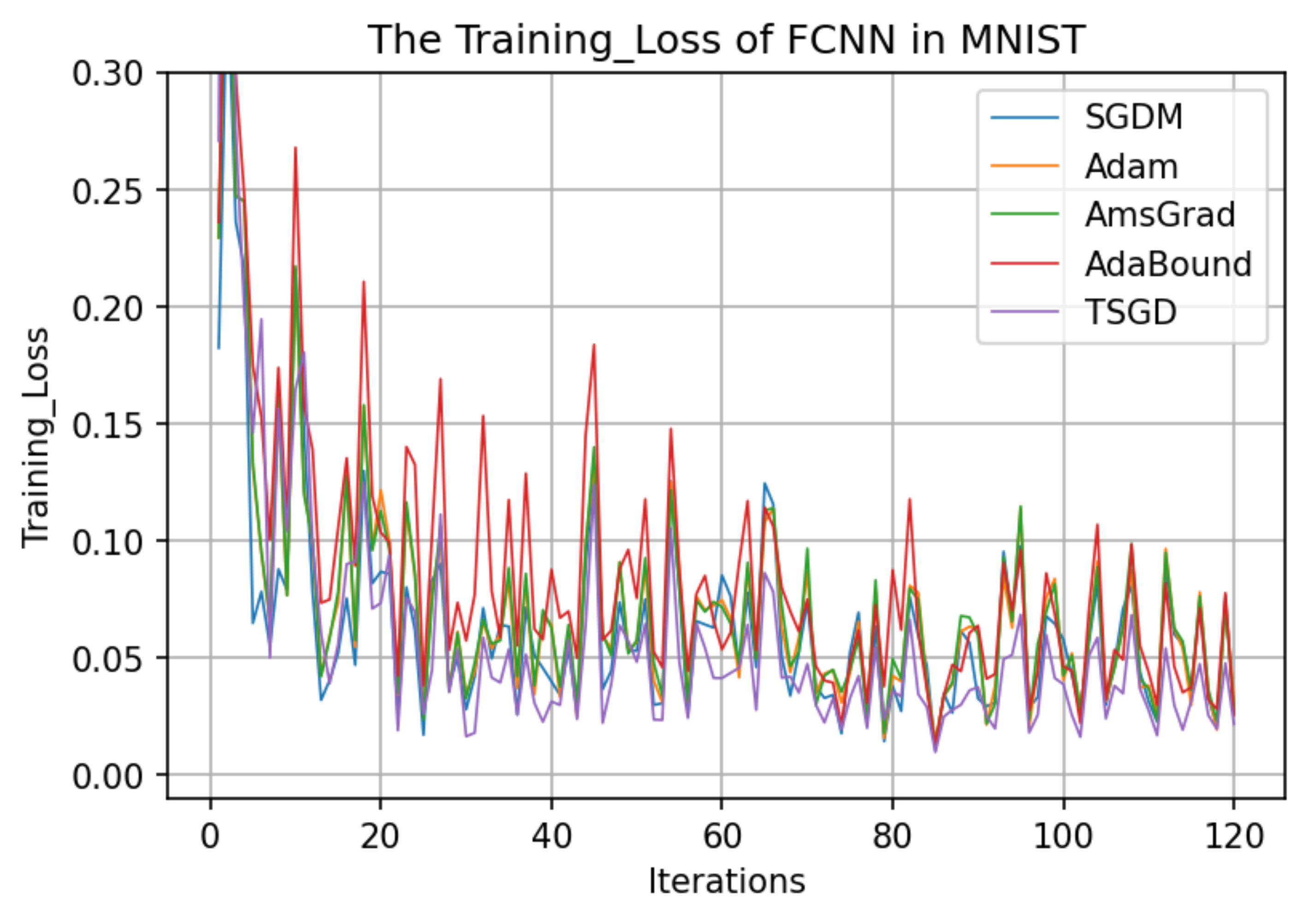}
\caption{The loss of MNIST-FCNN on training set}
\end{minipage}
\end{figure}

\begin{table}[!h]
\centering
\setcounter{table}{0}
\caption{The results of FCNN training MNIST without StepLR}
\begin{tabular}{lccccc}
\hline Optimizer/Result & NSGD & Adam & AmsGrad & AdaBound & TSGD \\
\hline Test Accuracy & $0.9813$ & $0.9813$ & $0.9811$ & $0.9797$ & $0.9829$ \\
Training Accuracy & $1.0000$ & $1.0000$ & $1.0000$ & $1.0000$ & $1.0000$ \\
Training Loss & $0.0105$ & $0.0144$ & $0.0129$ & $0.0126$ & $0.0098$ \\
\hline
\end{tabular}
\end{table}

\textbf {\fontsize{12pt}{0} \selectfont Cifar10 and ResNet18.} The Cifar10 data set\cite{krizhevsky2009learning} was collected by Hinton et al. It contains 60,000 color images in 10 categories, such as airplanes, cars, and birds. There are 50,000 training samples and 10,000 test samples . We use a more complex ResNet18 network. In this experiment, for the TSGD algorithm, we do not use the decreasing learning rate, and the learning rate is taken as a constant 0.05. For the hyperparameters of other algorithms, we use default values or empirical values, and we also do not use the interval of learning rate adjustment strategy(StepLR). The hyperparameter setting of the TSGD algorithm is: ($UR=0.05$, $LR=0.05$, $\beta=0.9$, $\gamma=1.000153$, epoch $=200$, dampening = $0$, batchSize $=128$, weight$_{-}$decay = 5$e$-4, seed = 1).

Figure 4 shows the accuracy of different algorithms on the test set. It can be seen from Figure 4, compared to the plain stochastic gradient descent, the accuracy, convergence speed, and stability of the TSGD algorithm have greatly improved. Compared with the adaptive gradient descent, the TSGD algorithm has a faster speed in the early stage of training, and even exceeds the convergence speed of the adaptive gradient descent. In the later stage of training, the accuracy has been greatly improved compared with other algorithms. Figure 5 shows the accuracy of different algorithms on the training set. The TSGD algorithm has the fastest convergence and the best accuracy. Figure 6 shows the loss of different algorithms on the training set. It can be seen that the loss of TSGD algorithm drops faster and the final loss is the smallest. We use the optimal value in the iterative process as the result of this experiments. The results are shown in Table 2.

\begin{figure}[!h]
\centering
\begin{minipage}[t]{0.48\textwidth}
\centering{}
\includegraphics[width=8cm]{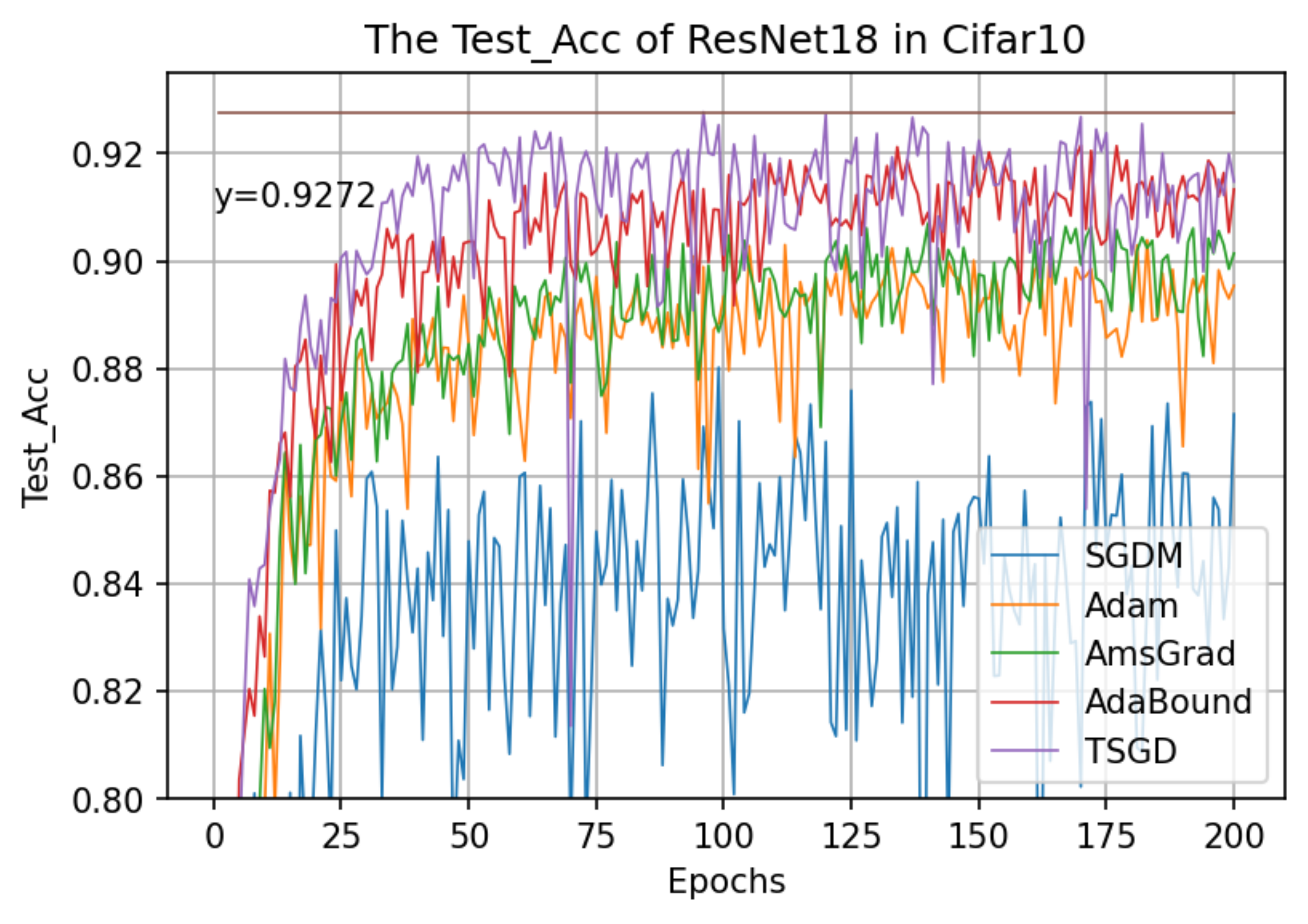}
\caption{The accuracy of Cifar10-ResNet18 on test set without StepLR}
\end{minipage}
\begin{minipage}[t]{0.48\textwidth}
\centering
\includegraphics[width=8cm]{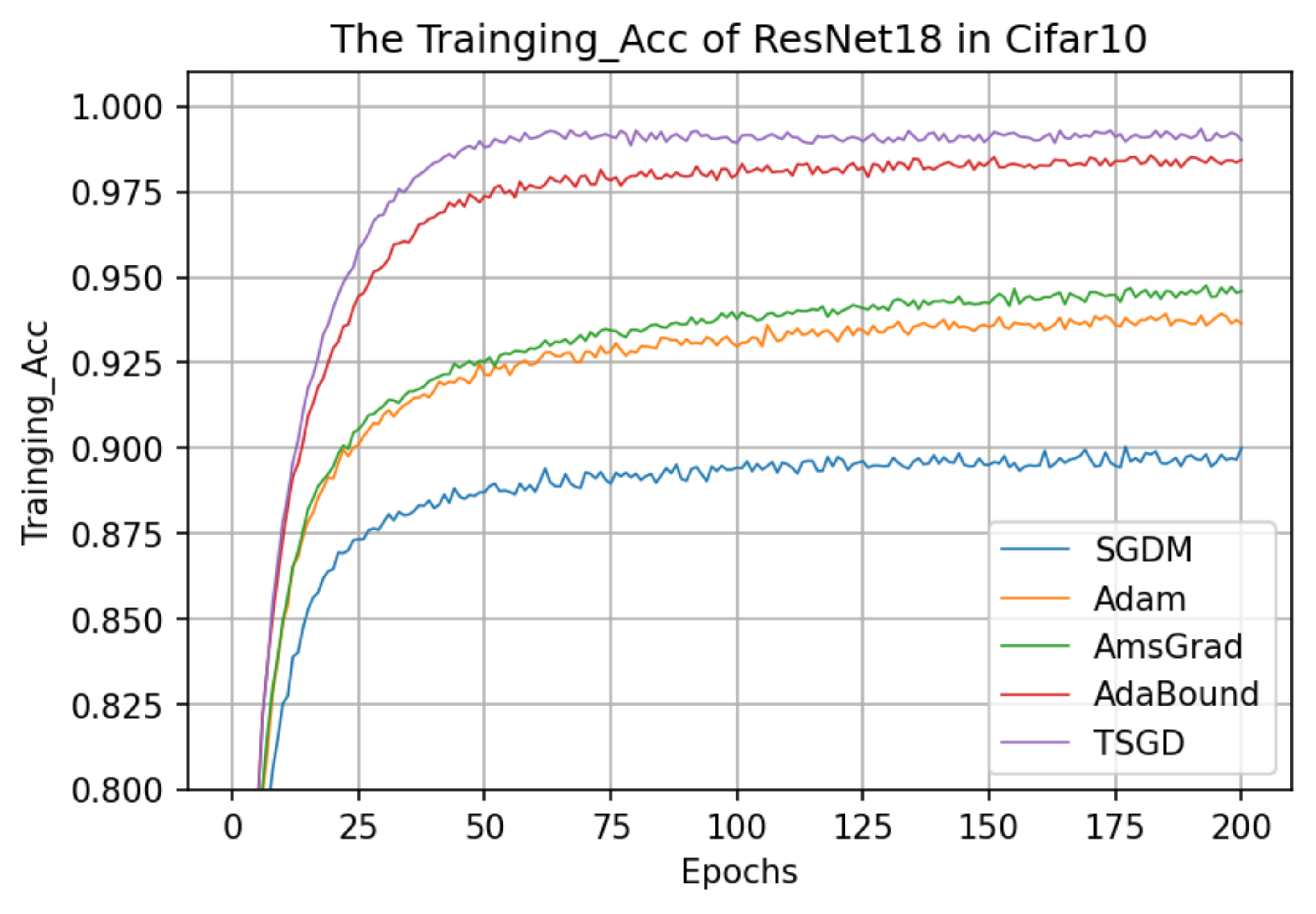}
\caption{The accuracy of Cifar10-ResNet18 on training set without StepLR}
\end{minipage}
\end{figure}

\begin{table}[!h]
\centering
\caption{The results of ResNet18 training Cifar 10 without StepLR}
\begin{tabular}{lccccc}
\hline Optimizer/Result & NSGD & Adam & AmsGrad & AdaBound & TSGD \\
\hline Test Accuracy & $0.8802$ & $0.9039$ & $0.9070$ & $0.9215$ & $0.9277$ \\
Training Accuracy & $0.9004$ & $0.9392$ & $0.9476$ & $0.9856$ & $0.9980$ \\
Training Loss & $0.2961$ & $0.1776$ & $0.1534$ & $0.0447$ & $0.0218$ \\
\hline
\end{tabular}
\end{table}

Under the above experimental conditions, we use a decreasing learning rate for the TSGD algorithm. For other algorithms, we use the interval of learning rate adjustment strategy(StepLR), and divide the learning rate by 10 after the 150-$th$ iteration. Figure 7 shows the accuracy of different algorithms on the test set. It can be seen that when the learning rate is divide by 10 , the accuracy of the other algorithms has risen sharply. The Adabound algorithm has better result than other algorithms. However, the TSGD algorithm has better performance, and its convergence speed even exceeds the adaptive gradient descent in the early stage of training. The highest accuracy is achieved in the later stage of training. We use the optimal value of the training process as the result of this experiment, and the results are shown in Table 3.

\begin{figure}[!h]
\centering
\begin{minipage}[t]{0.48\textwidth}
\centering
\includegraphics[width=8cm]{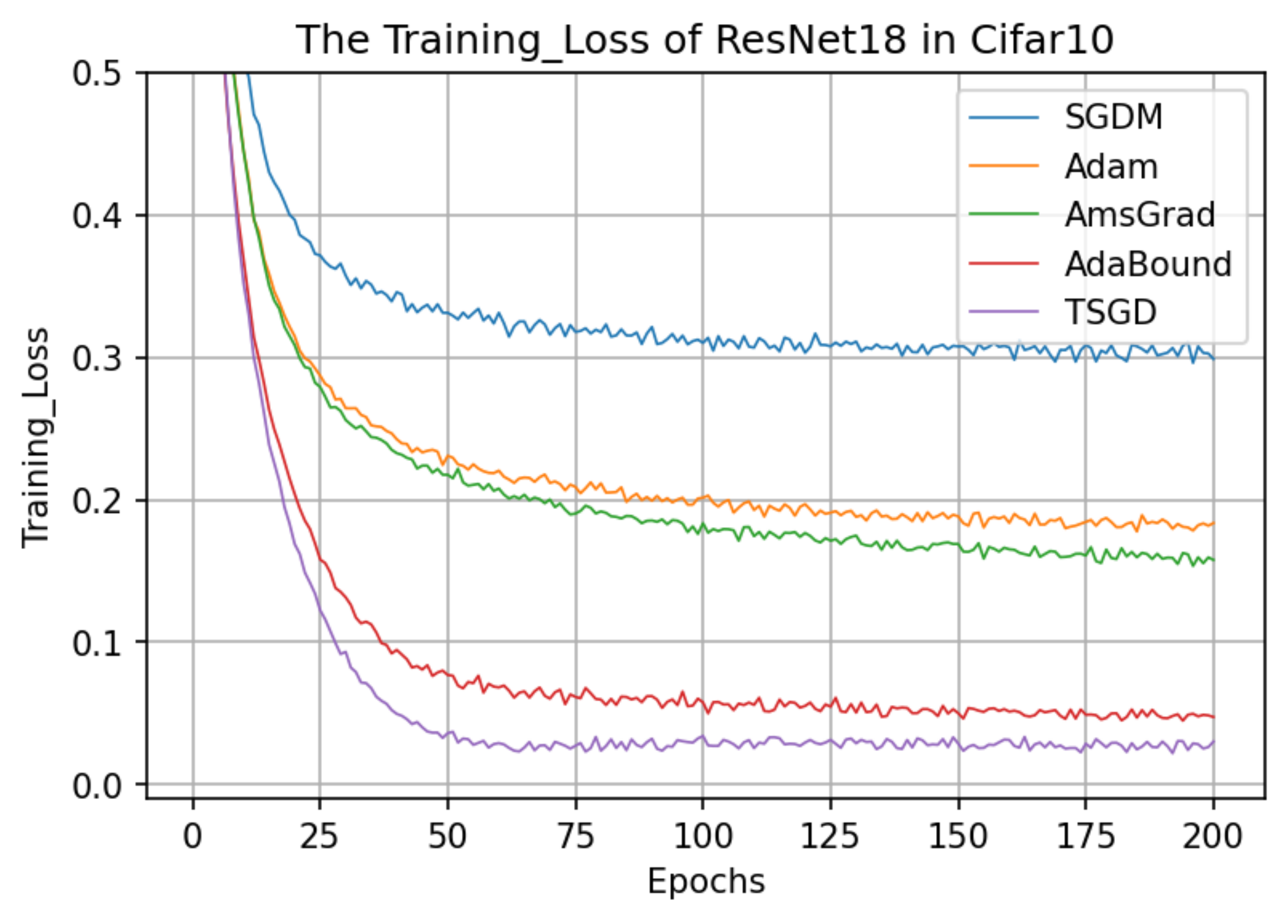}
\caption{The loss of Cifar10-ResNet18 on training set without StepLR}
\end{minipage}
\begin{minipage}[t]{0.48\textwidth}
\centering
\includegraphics[width=8cm]{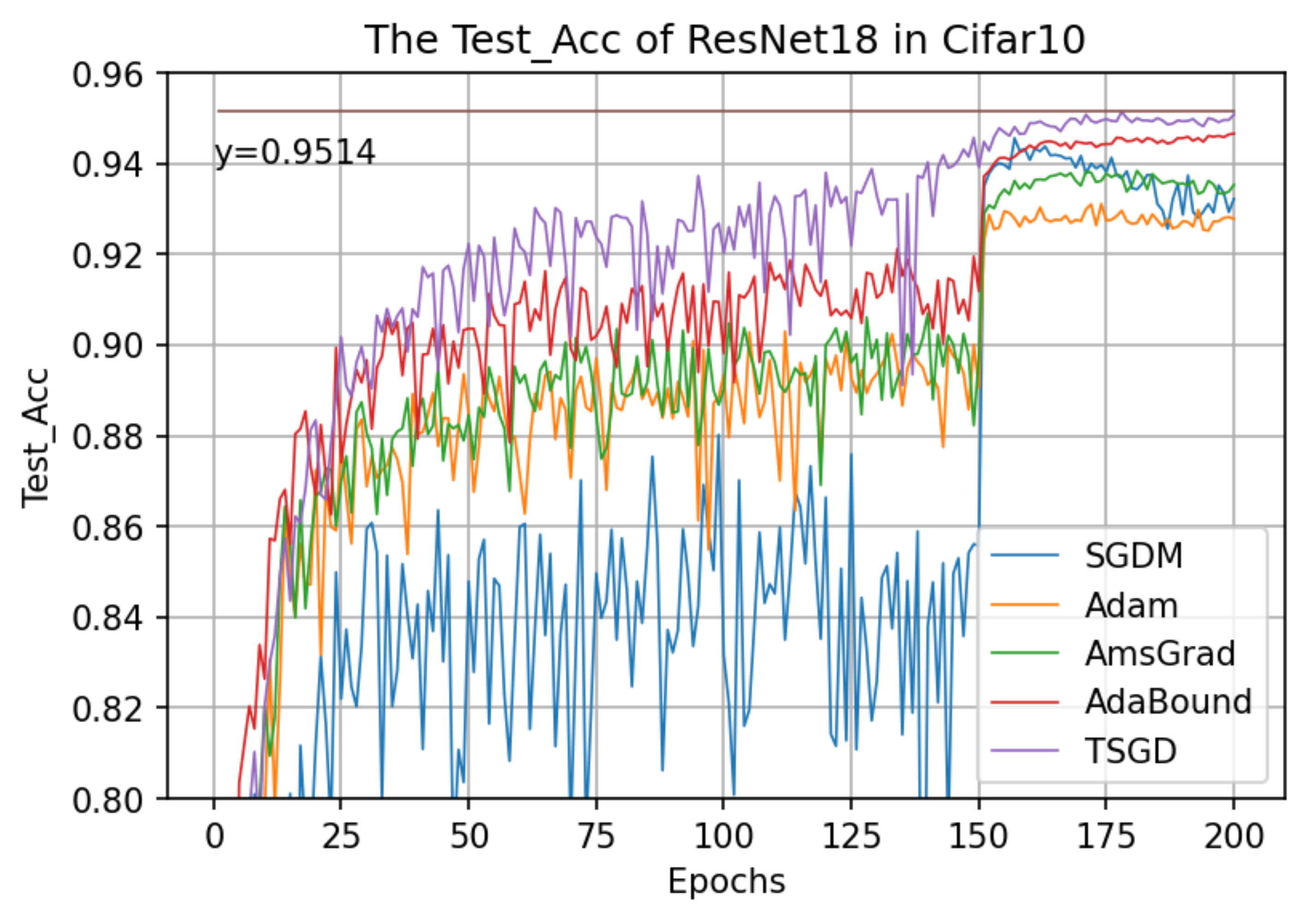}
\caption{The accuracy of Cifar10-ResNet18 on test set with StepLR}
\end{minipage}
\end{figure}

\begin{table}[!h]
\centering
\caption{The results of ResNet18 training Cifar 10 with StepLR}
\begin{tabular}{lccccc}
\hline Optimizer/Result & NSGD & Adam & AmsGrad & AdaBound & TSGD \\
\hline Test Accuracy & $0.9456$ & $0.9311$ & $0.9384$ & $0.9466$ & $0.9514$ \\
Training Accuracy & $0.9916$ & $0.9936$ & $0.9976$ & $0.9999$ & $1.0000$ \\
Training Loss & $0.0270$ & $0.0206$ & $0.0116$ & $0.0014$ & $0.0015$ \\
\hline
\end{tabular}
\end{table}
Similarly, we do the experiment on the cifar100 data set to compare different algorithms. Cifar100 is similar to Cifar10, but Cifar100 has 10 categories. Each category has 10 subcategories and total of 100 categories. Each category contains 600 pictures. The 500 pictures are used to train and 100 pictures are used to test. The neural network model we still use ResNet18. In this experiment, for the TSGD algorithm, we do not use the decreasing learning rate, and the learning rate is taken as a constant 0.05.  For other algorithms, we also do not use the interval of learning rate adjustment strategy(StepLR). And the hyperparameters, we use default values or empirical values. The hyperparameter setting of the TSGD algorithm is: ($UR=0.05$, $LR=0.05$, $\beta=0.9$, $\gamma=1.000153$, epoch $=200$, dampening = $0$, batchSize $=128$, weight$_{-}$decay = 5$e$-4, seed = 1).

Figure 8 shows the accuracy of different algorithms on the test set. It can be seen that TSGD algorithm has faster speed in the early stage of training. There are two sudden changes in accuracy curve, which may fall into the local optimum. But with learning rate decreasing, the accuracy gradually rises, and finally stabilizes. Figure 9 shows the accuracy of the training set. It can be seen that the TSGD algorithm on the training set has fast speed and high accuracy. Figure 10 shows the loss on the training set. It can be seen that the loss of the TSGD algorithm drops faster, and finally has a smaller loss. We use the optimal value of the training process as the result of this experiment, and the results are shown in Table 4. 

\begin{figure}[!h]
\centering
\begin{minipage}[t]{0.48\textwidth}
\centering
\includegraphics[width=8cm]{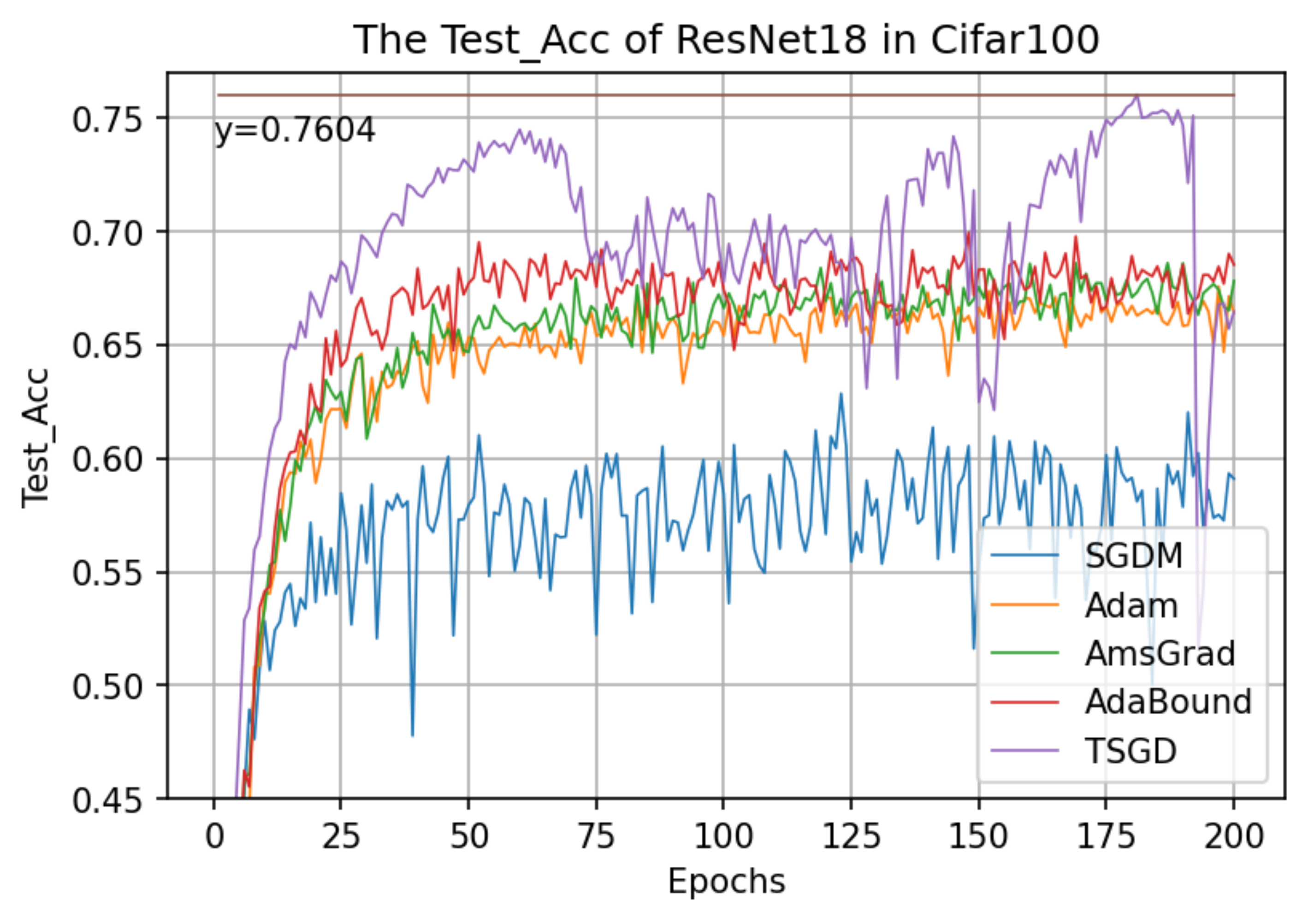}
\caption{The loss of Cifar100-ResNet18 on training set without StepLR}
\end{minipage}
\begin{minipage}[t]{0.48\textwidth}
\centering
\includegraphics[width=8cm]{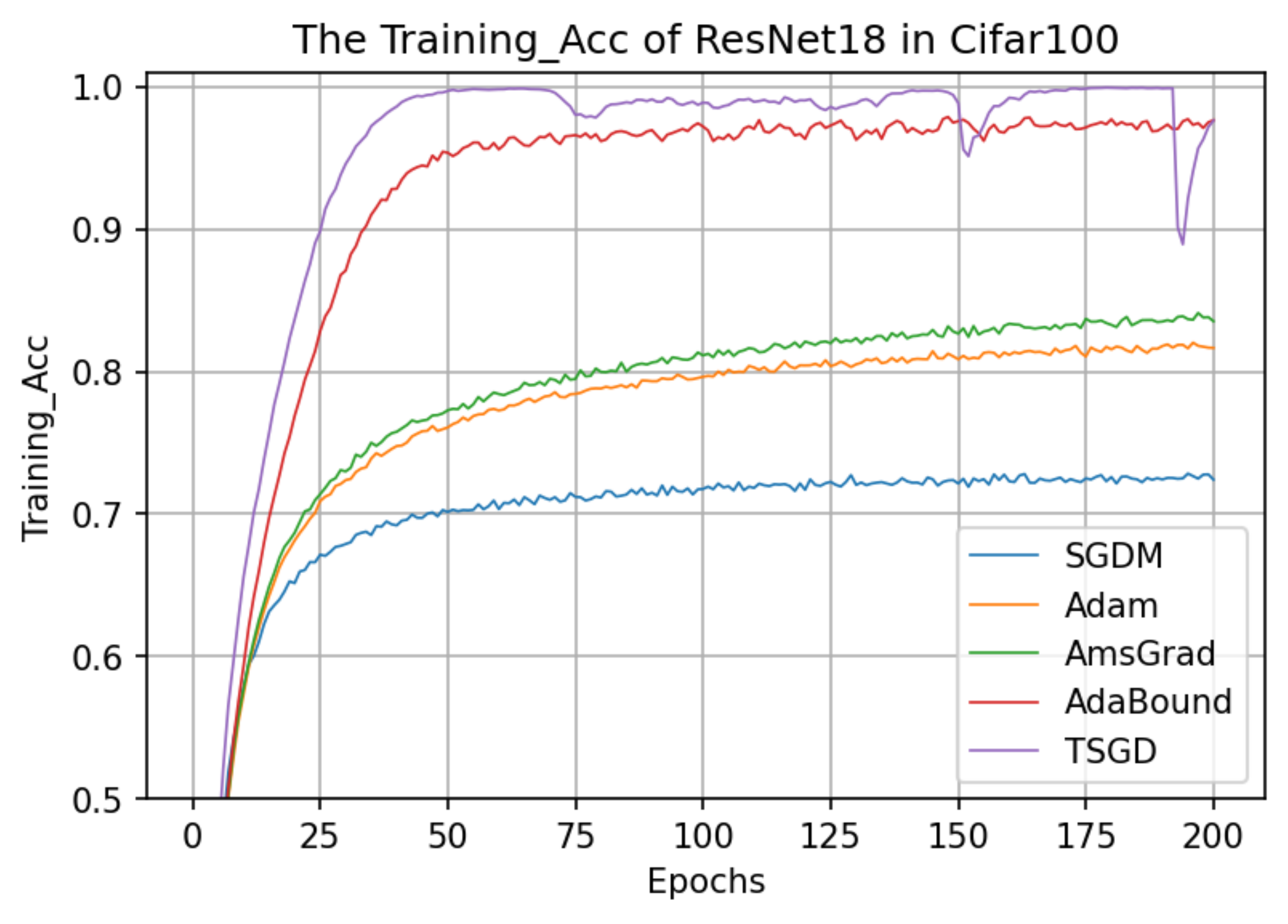}
\caption{The accuracy of Cifar100-ResNet18 on test set without StepLR}
\end{minipage}
\end{figure}

\begin{table}[!h]
\centering
\caption{The results of ResNet18 training Cifar100 witout StepLR}  
\begin{tabular}{lccccc}
\hline Optimizer/Result & NSGD & Adam & AmsGrad & AdaBound & TSGD \\
\hline Test Accuracy & $0.6286$ & $0.6762$ & $0.6862$ & $0.6995$ & $0.7604$ \\
Training Accuracy & $0.7283$ & $0.8203$ & $0.8412$ & $0.9790$ & $0.9995$ \\
Training Loss & $0.9218$ & $0.5840$ & $0.5243$ & $0.0817$ & $0.0075$ \\
\hline
\end{tabular}
\end{table}

Under the above experimental conditions, we use a decreasing learning rate for the TSGD algorithm, and for other algorithms we use the interval of learning rate adjustment strategy(StepLR). we divide the learning rate by 10 after the 150-$th$ iteration. Figure 11 shows the accuracy of different algorithms on the test set. It can be seen that the final accuracy of the TSGD algorithm that uses a decreasing learning rate is highest. We use the optimal value of the training process as the result of this experiment, and the results are shown in Table 5. 

\begin{figure}[!h]
\centering
\begin{minipage}[t]{0.48\textwidth}
\centering
\includegraphics[width=8cm]{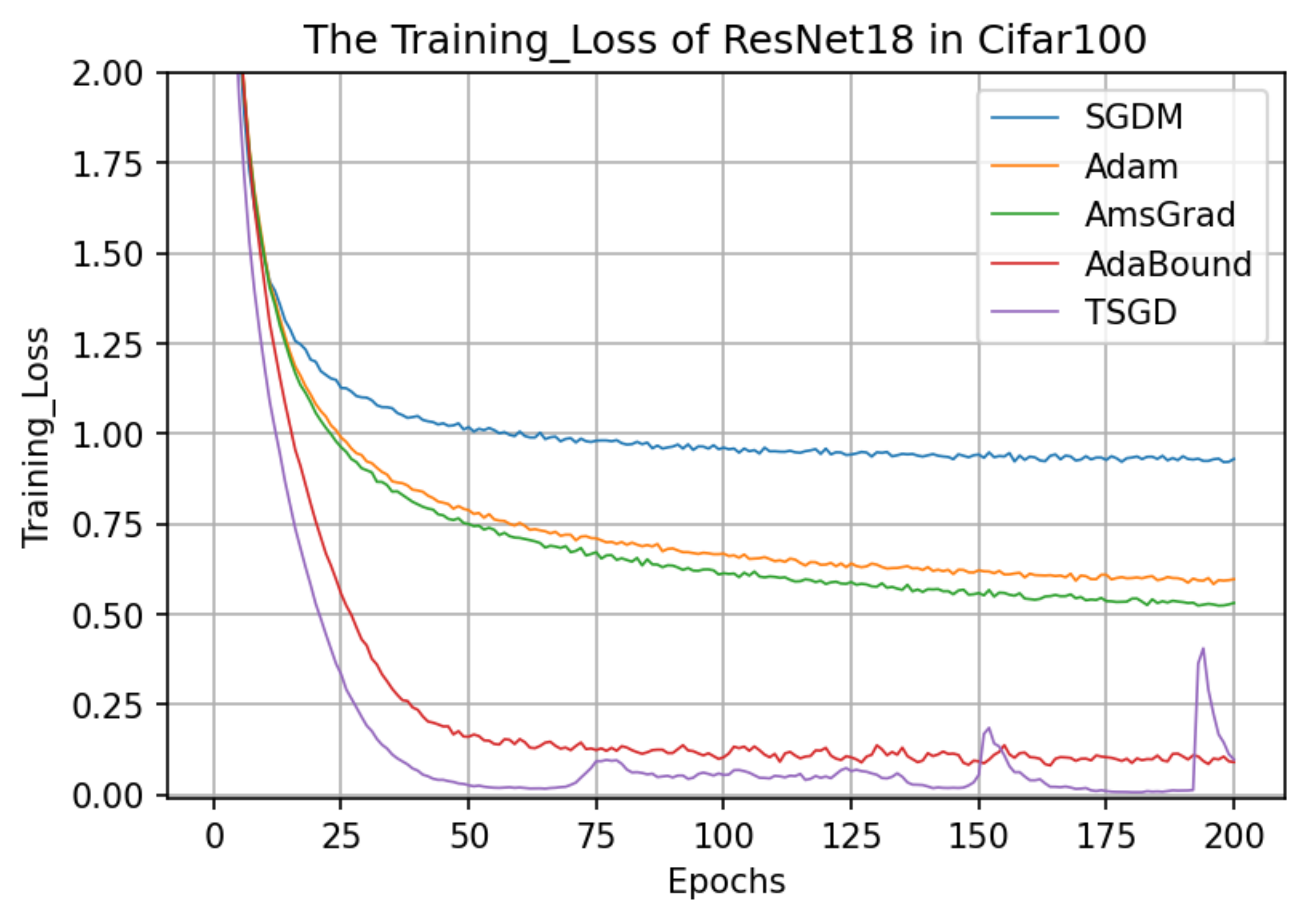}
\caption{The loss of Cifar100-ResNet18 on training set without StepLR}
\end{minipage}
\begin{minipage}[t]{0.48\textwidth}
\centering
\includegraphics[width=8cm]{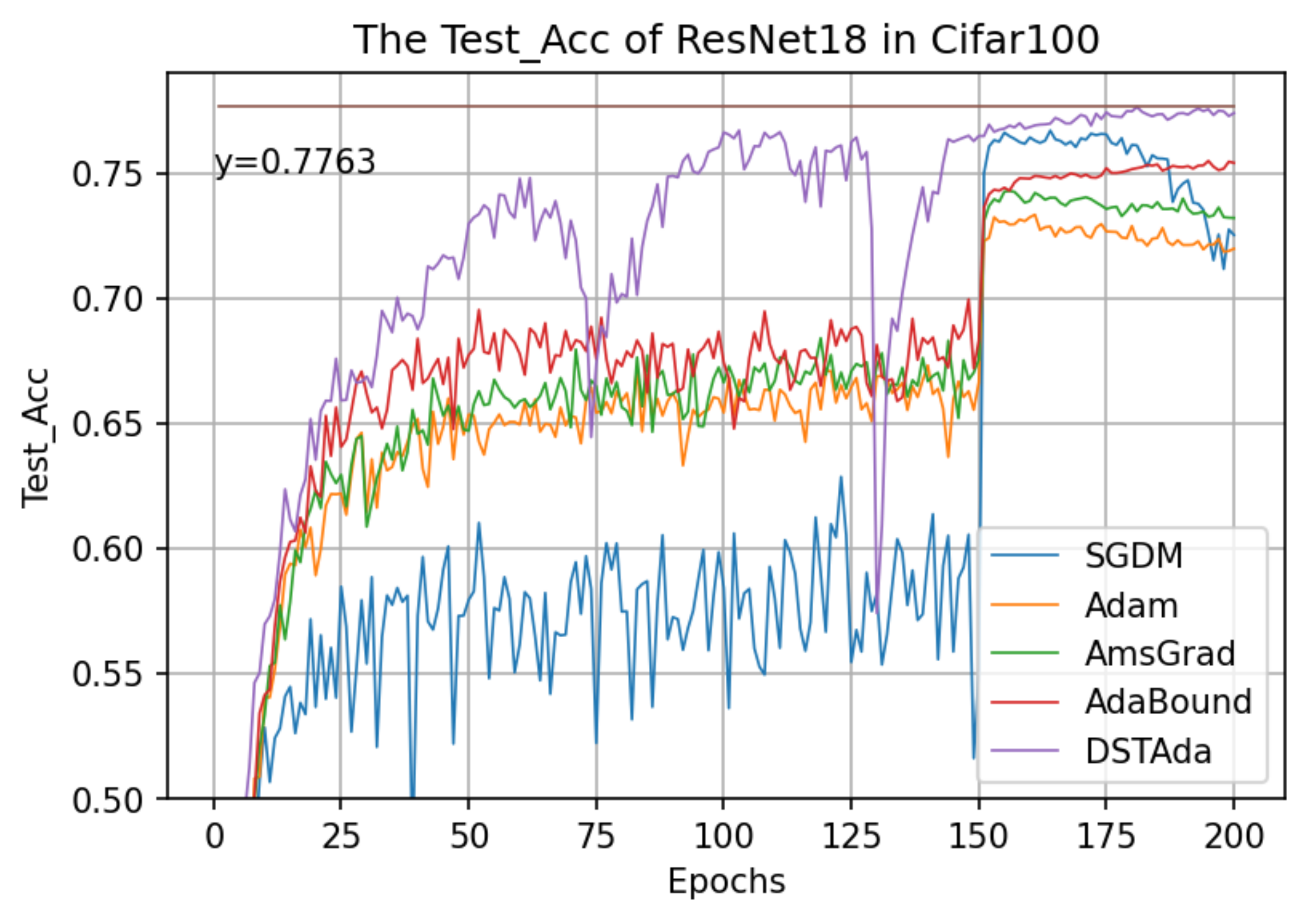}
\caption{The accuracy of Cifar100-ResNet18 on test set with StepLR}
\end{minipage}
\end{figure}

\begin{table}[!h]
\centering
\caption{The results of ResNet18 training Cifar100 with StepLR}  
\begin{tabular}{lccccc}
\hline Optimizer/Result & NSGD & Adam & AmsGrad & AdaBound & TSGD \\
\hline Test Accuracy & $0.7669$ & $0.7333$ & $0.7428$ & $0.7548$ & $0.7763$ \\
Training Accuracy & $0.9942$ & $0.9833$ & $0.9906$ & $0.9998$ & $0.9998$ \\
Training Loss & $0.0391$ & $0.0743$ & $0.0199$ & $0.0052$ & $0.0064$ \\
\hline
\end{tabular}
\end{table}

In this section, we use experiments to verify the performance of the TSGD algorithm proposed in this paper. From the experiments, it can be seen that TSGD achieves the best results in terms of convergence speed and accuracy. Compared with the plain stochastic gradient descent, the convergence speed, stability, and accuracy of TSGD algorithm have been greatly improved. Compared with the adaptive gradient descent, TSGD does not need to calculate the second moment, reducing the computational complexity. In particular, the convergence speed even exceeds the convergence speed of the adaptive gradient descent. In general, the TSGD algorithm has better performance.

\section{Extensions}
\label{S:6}

\textbf {\fontsize{12pt}{0} \selectfont Sensitivity of hyperparameters.} In this section, we verify the sensitivity of the hyperparameters UR and LR of the TSGD algorithm. We select different upper and lower learning rate for testing. The data set is cifar10, and the neural network model is ResNet18. The hyperparameter setting of the TSGD algorithm is: ($UR=0.1$, $LR=0.005$, $\beta=0.9$, $\gamma=1.000153$, epoch $=200$, dampening = $0$, batchSize $=128$, weight$_{-}$decay = 5$e$-4, seed = 1).

When testing different upper learning rate $UR$, we fix the lower learning rate $LR=0.005$. When testing the lower learning rate $LR$, we fix the upper learning rate $UR = 0.1$. Figure 12 shows the accuracy of different upper learning rate on the test set. Figure 13 shows the accuracy of different lower learning rate on the test set. It can be seen that the TSGD algorithm does not have a very strong dependence on the upper and lower learning rate. The final accuracy also has a smaller gap. For this training model, the upper learning rate  $UR = 0.1$ and lower learning rate $LR = 0.005$ of the TSGD algorithm has a better convergence result, which are two very good recommendations. 

\begin{figure}[!h]
\centering
\begin{minipage}[t]{0.48\textwidth}
\centering
\includegraphics[width=8cm]{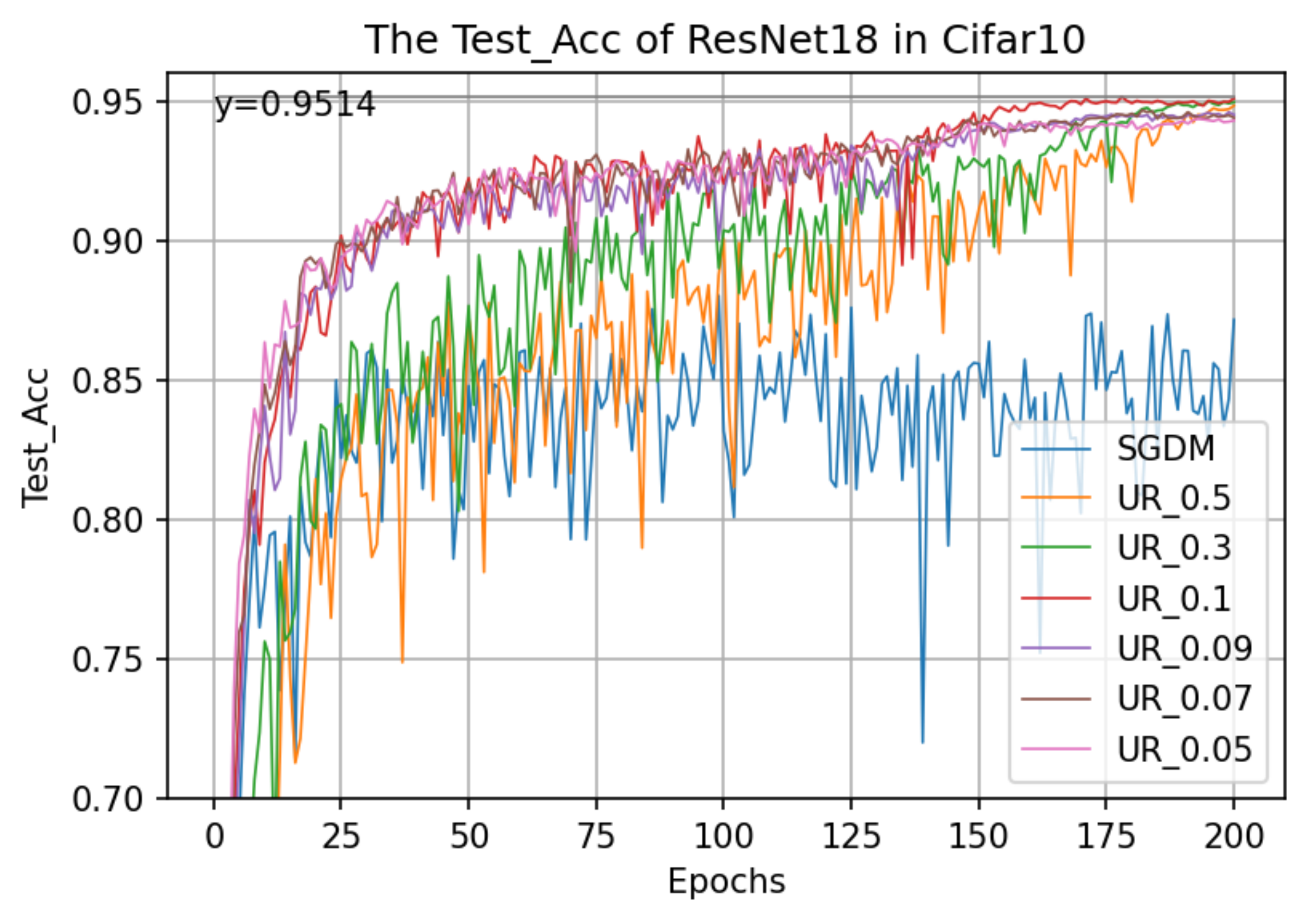}
\caption{The accuracy of Cifar10-ResNet18 on test set for different $UR$}
\end{minipage}
\begin{minipage}[t]{0.48\textwidth}
\centering
\includegraphics[width=8cm]{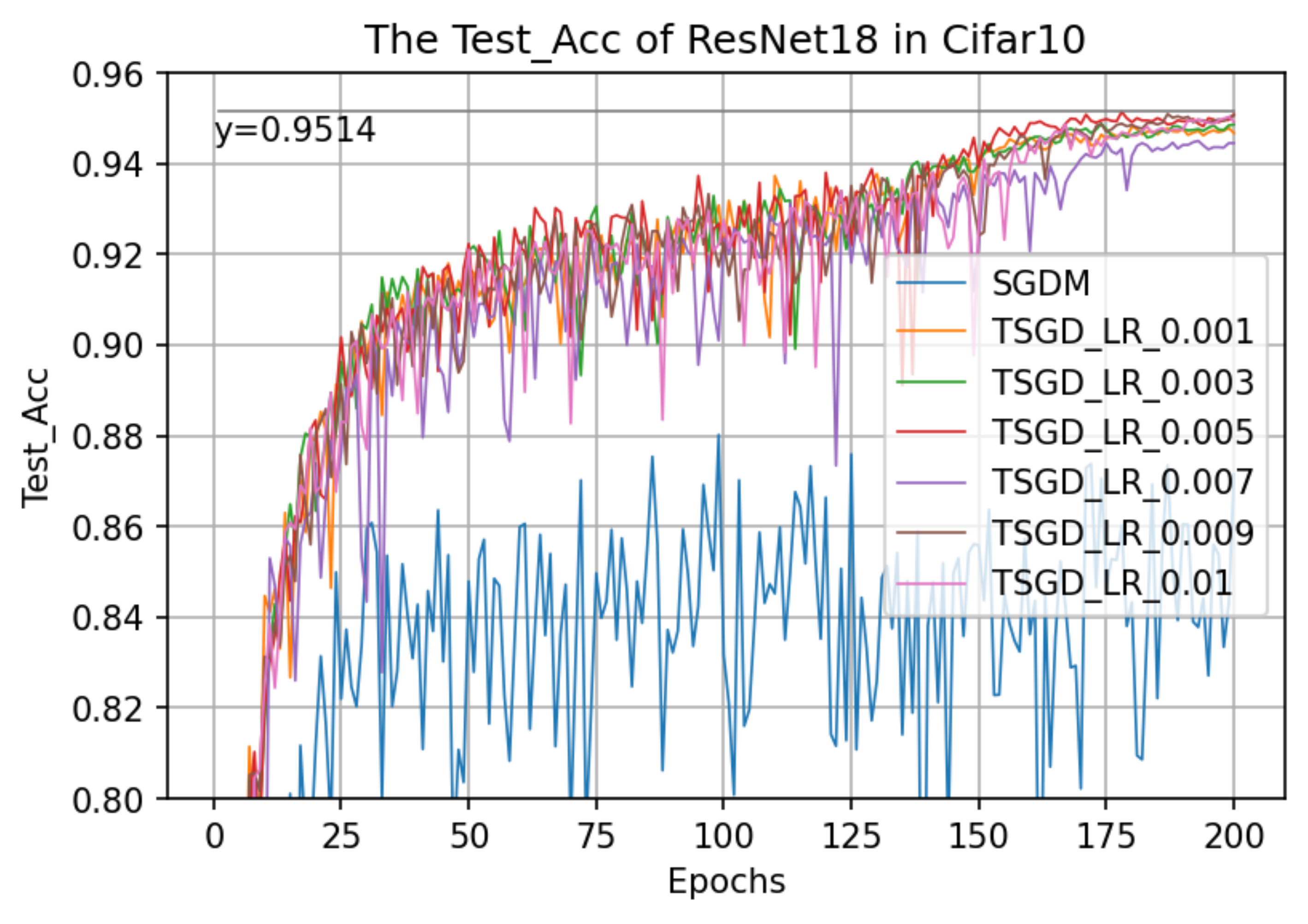}
\caption{The accuracy of Cifar10-ResNet18 on test set for different $LR$}
\end{minipage}
\end{figure}

\textbf {\fontsize{12pt}{0} \selectfont Selection of hyperparameter $\gamma $}. In the TSGD algorithm, compared to the heavy-ball momentum gradient descent algorithm, we introduce more hyperparameters $UR$, $LR$, $\gamma $. In the above experiment, we can see that the TSGD algorithm does not have a very strong dependence on the upper and lower learning rate. Therefore, here we only discuss the choice of $\gamma $. $\gamma $ as a conversion factor, it controls the moment of transition from the momentum gradient descent to the plain stochastic gradient descent. The main formula for the transition is: ${\hat m_t} = \left( {{m_t} - {g_t}} \right)1/{\gamma ^t} + {g_t}$. When the magnitude of $1/{\gamma ^t}$ decreases by  $k = {10^{ - 2}}$( the magnitude of the difference between ${m_t}$ and  ${g_t}$ is $k$), we think that the transition has been achieved. During the experiment, $T$  denotes the total number of iterations, $sampleSize$ denotes the total number of training samples, $batchSize$ denotes the size of each batch. According to the above variables, we can calculate the specific value of $\gamma $ at a certain transition point. For example: we choose the first 3/8 of the total number of iterations $T$ for momentum gradient descent training, and the last 5/8 for plain stochastic gradient descent training. The total number of samples is $sampleSize = 50000$, the batch size is $batchSize = 128$, the epochs is $epoch = 200$, and the magnitude is $k = 0.01$ (in fact, through the solving process of $\gamma $, we can know that the value of $k$ has little effect on $\gamma $). Then we have $T = sampleSize/batchSize*epoch$, and take $T = 80000$. Since when the magnitude of $1/{\gamma ^t}$ is reduced by $k = 0.01$, we think that a transition has been achieved. so there is $k = 1/{\gamma ^{T3/8}}$, and $\gamma  = 1.000153$. Obviously, the selection of hyperparameters $\gamma $ is expected and controllable for the experimental results, and does not require empirical selection or repeated experimental selection.

\begin{figure}[!h]
\centering
\begin{minipage}[t]{0.48\textwidth}
\centering
\includegraphics[width=8cm]{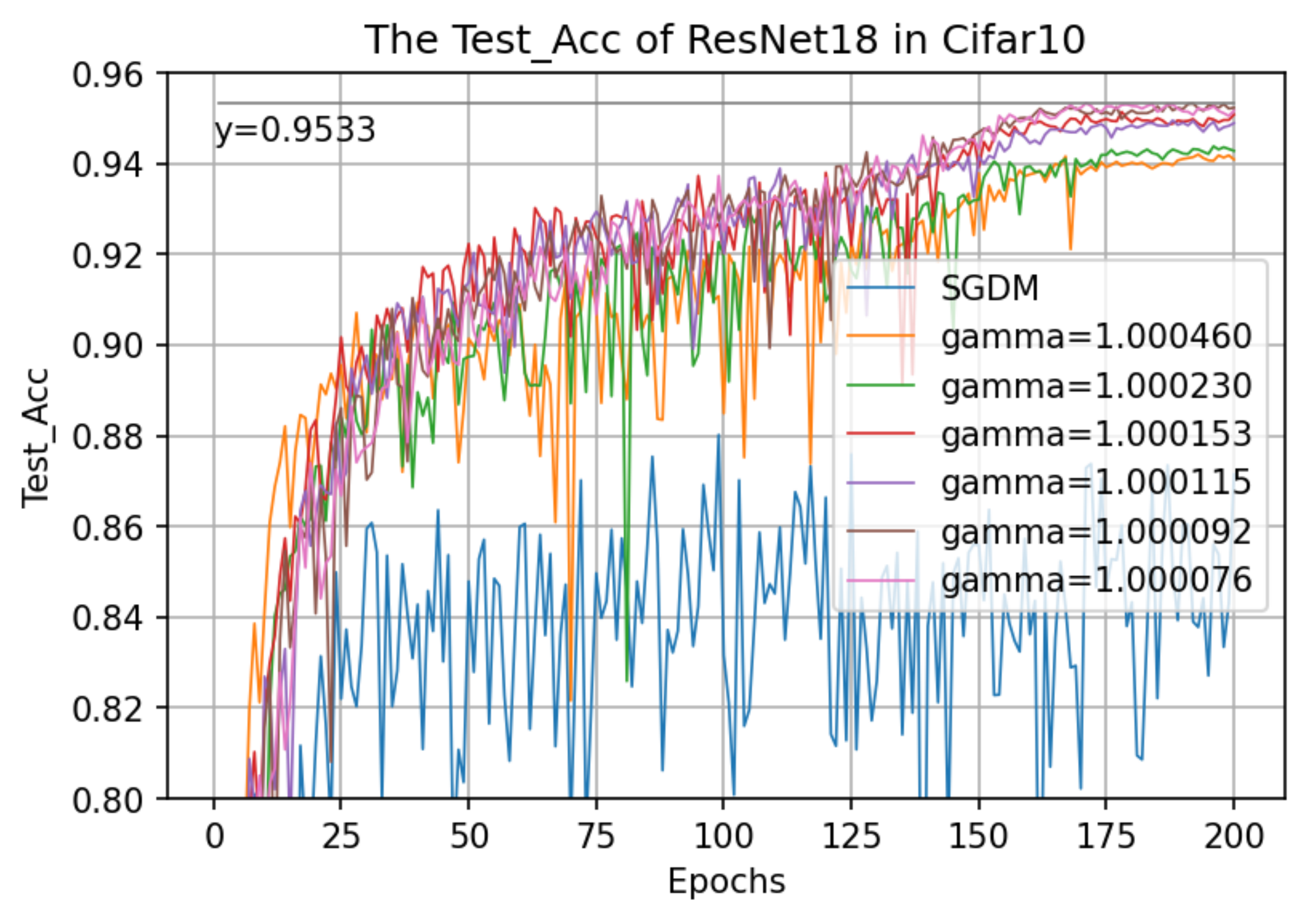}
\caption{The accuracy of Cifar10-ResNet18 on test set for different $\gamma$}

\end{minipage}
\begin{minipage}[t]{0.48\textwidth}
\centering
\includegraphics[width=8cm]{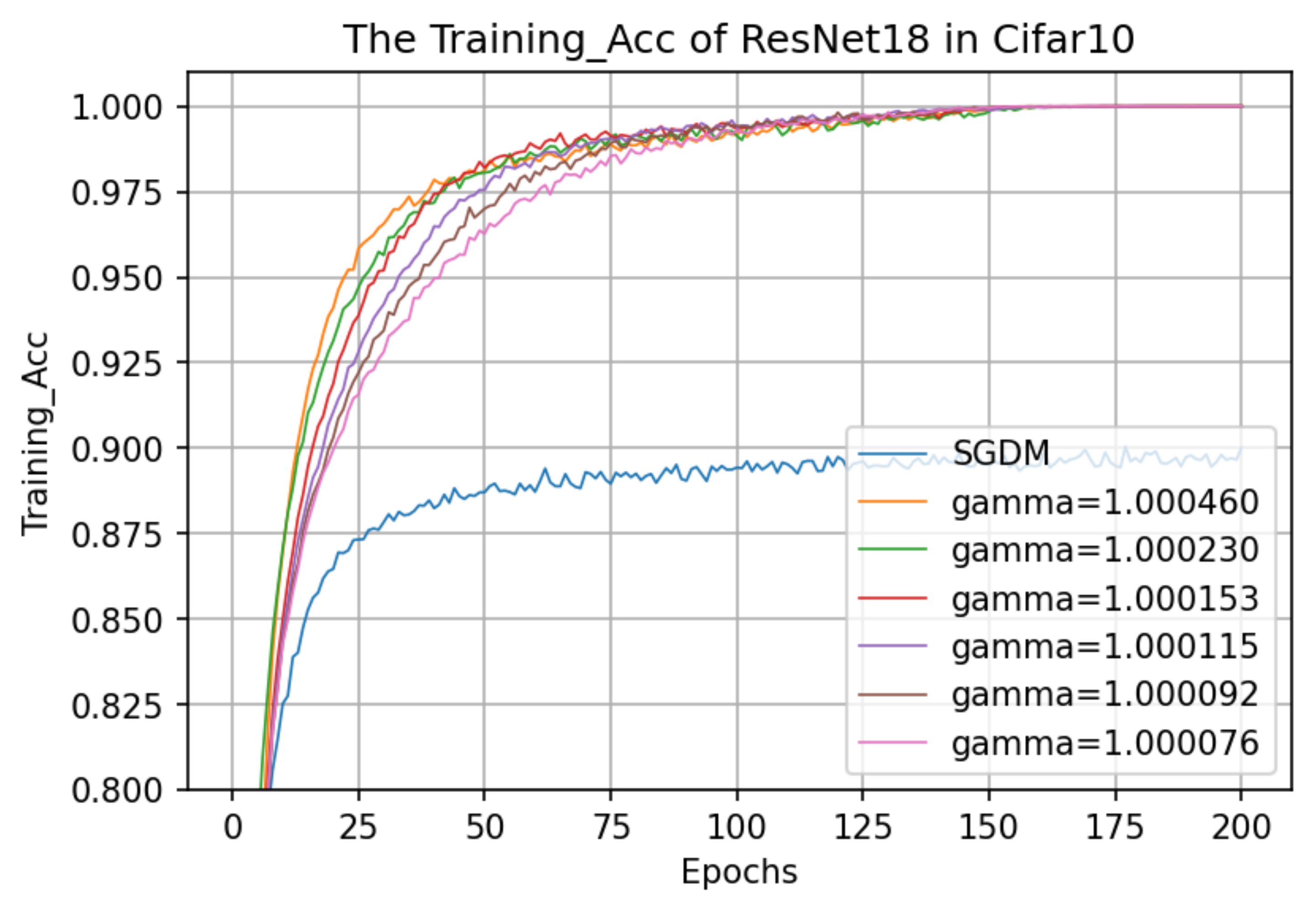}
\caption{The accuracy of Cifar10-ResNet18 on training set for different $\gamma$}
\end{minipage}
\end{figure}

\begin{figure}[!h]
\centering
\begin{minipage}[t]{0.48\textwidth}
\centering
\includegraphics[width=8cm]{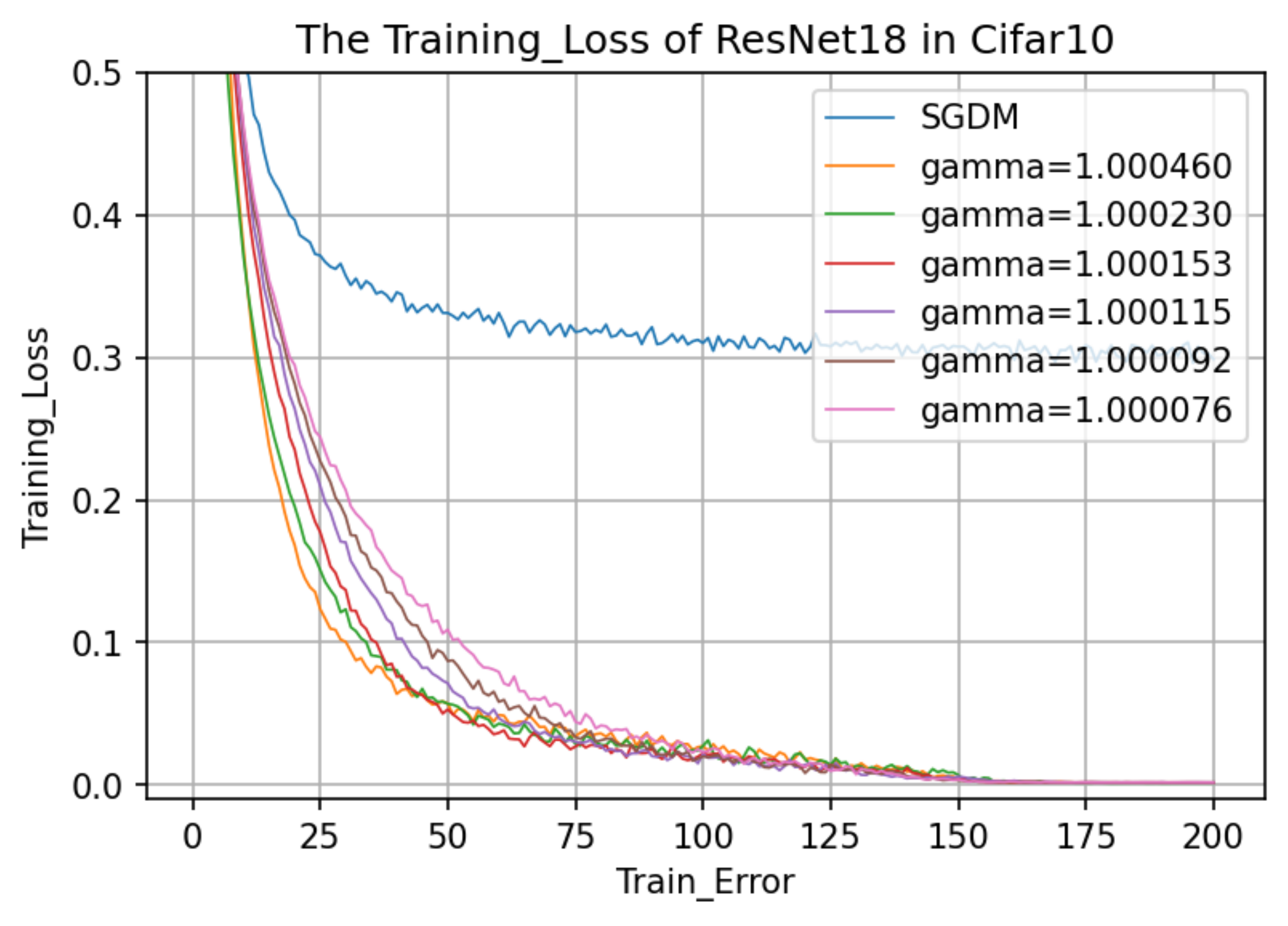}
\caption{The loss of Cifar10-ResNet18 on training set for different $\gamma$}

\end{minipage}
\begin{minipage}[t]{0.48\textwidth}
\centering
\includegraphics[width=8cm]{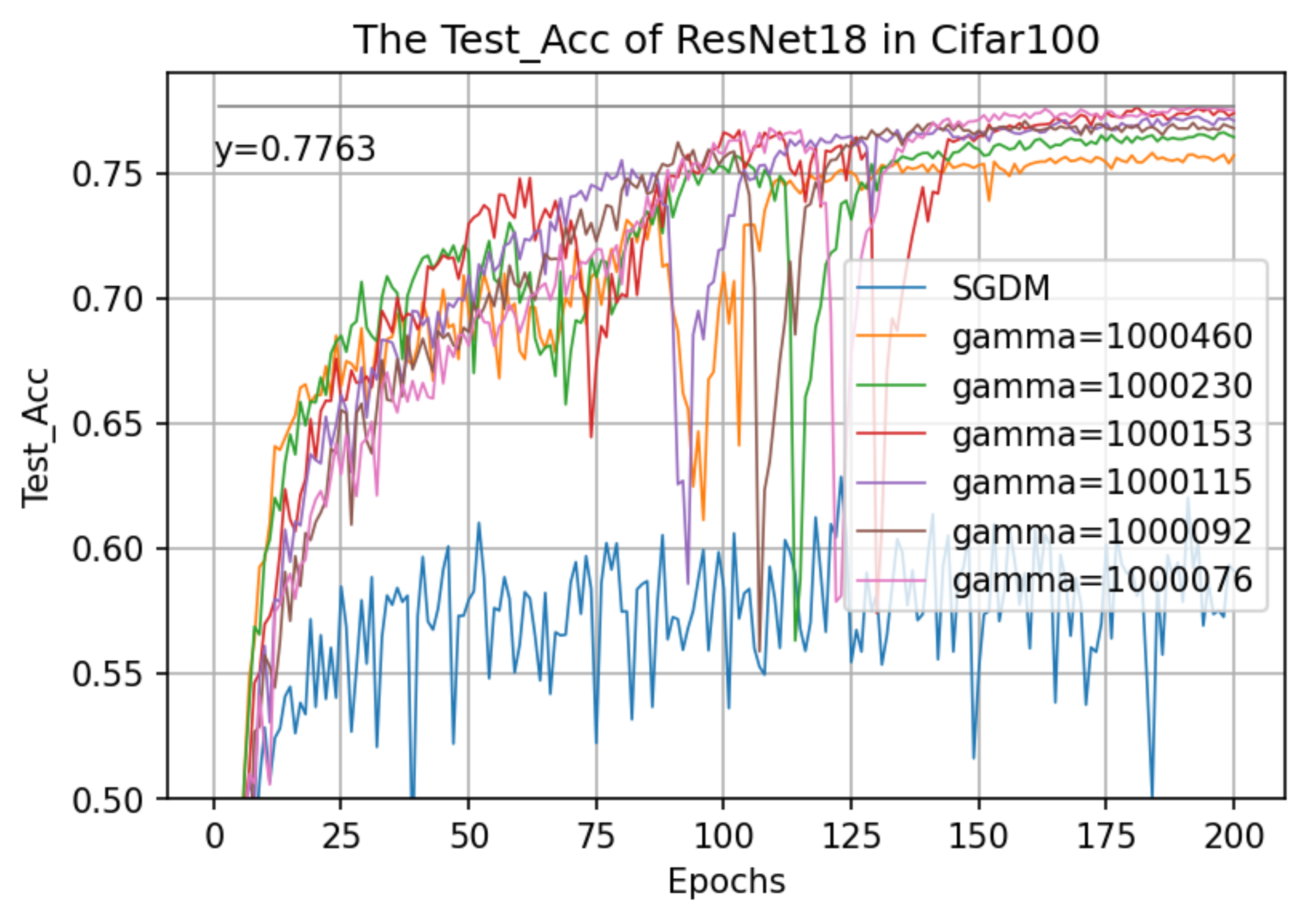}
\caption{The accuracy of Cifar100-ResNet18 on test set for different $\gamma$}
\end{minipage}
\end{figure}

\begin{table}[!h]
\centering
\caption{The values of $\gamma$ with respect to different transition points} 
\begin{tabular}{llllllll}
\hline Name/No & 1 & 2 & 3 & 4 & 5 & 6 & \\
\hline $\mathrm{t}$ & 10000 & 20000 & 30000 & 40000 & 50000 & 60000 & \\
$\mathrm{t} / \mathrm{T}$ & $1 / 8$ & $1 / 4$ & $3 / 8$ & $1 / 2$ & $5 / 8$ & $3 / 4$ & \\
$\gamma$ & $1.000460$ & $1.000230$ & $1.000153$ & $1.000115$ & $1.000092$ & $1.000076$ \\
\hline
\end{tabular}
\end{table}

We use ResNet18 to train the Cifar10 data set. In Figures 14, 15, and 16, it shows the accuracy of the test set and training set, and the loss value of the training set for different values of $\gamma $. We use ResNet18 to train the Cifar100 data set. In Figure 17, it shows the accuracy of different values of $\gamma $ on the test set. For larger values of $\gamma $, the convergence effect of TSGD is not so ideal. It can be seen from Figures 14 and 17, that the accuracy of the SGDM algorithm is no longer increased when the SGDM algorithm is applied to the plateau about 50 epochs. Therefore, empirical selection of transition points after the plateau usually has a better result. For the remaining  values of $\gamma $, the TSGD algorithm can converge very well, and the final gap is also small. For the accuracy and loss value of training set, the final gap is relatively small, and the final curves almost overlap. In Table 6, the value of $\gamma $ corresponding to different transition points are given. 

\section{Conclusion}
\label{S:7}
In this paper, we combine the advantages of the momentum stochastic gradient descent with fast training speed and the plain stochastic gradient descent with high accuracy, and propose the scaling transition from momentum stochastic gradient descent to plain stochastic gradient descent method. At the same time, we use a decreasing learning rate, eliminating the need for complicated procedures of the interval of learning rate adjustment(StepLR). They make TSGD algorithm has faster convergence speed and higher accuracy. The experimental results show that the TSGD algorithm has good performance in terms of convergence speed and generalization ability.

\section*{Acknowledgements}
This work was funded in part by National Key R\&D Program of China (No.2020YFA0714102), and in part by the Fundamental Research Funds for the Central Universities of China (No.2412020F Z024).

\newpage

\bibliographystyle{model1-num-names}

\bibliography{source.bib}


%
%
%
%
%










\end{document}